\documentclass[a4paper,fleqn]{cas-dc}

\usepackage[numbers]{natbib}
\usepackage{subfig}
\usepackage{float}

\begin{document}
\def\floatpagepagefraction{1}
\def\textpagefraction{.01}
\shorttitle{Fangze~Fu et~al. Knowledge-Based Systems}
\shortauthors{Fu et~al.}

\title [mode = title]{SDR-GNN: Spectral Domain Reconstruction Graph Neural Network for Incomplete Multimodal Learning in Conversational Emotion Recognition}

\tnotetext[1]{Our code is publicly available at \href{https://github.com/fufangze/SDR-GNN}{https://github.com/fufangze/SDR-GNN}.}

\author[1]{Fangze Fu}

\author[1]{Wei Ai}

\author[1]{Fan Yang}

\author[1]{Yuntao Shou}

\author[1]{Tao Meng}
\cormark[1]

\author[2]{Keqin Li}

\cortext[1]{Corresponding author}

\address[1]{organization={College of Computer and Mathematics, Central South University of Forestry and Technology},
                postcode={410004},
                city={ Hunan, Changsha},
                country={China}}

\address[2]{organization={Department of Computer Science, State University of New York}, 
                city={New Paltz, New York},
                postcode={12561},
                country={USA}}

\begin{abstract}
Multimodal Emotion Recognition in Conversations (MERC) aims to classify utterance emotions using textual, auditory, and visual modal features. Most existing MERC methods assume each utterance has complete modalities, overlooking the common issue of incomplete modalities in real-world scenarios. Recently, graph neural networks (GNNs) have achieved notable results in Incomplete Multimodal Emotion Recognition in Conversations (IMERC). However, traditional GNNs focus on binary relationships between nodes, limiting their ability to capture more complex, higher-order information. Moreover, repeated message passing can cause over-smoothing, reducing their capacity to preserve essential high-frequency details.
To address these issues, we propose a Spectral Domain Reconstruction Graph Neural Network (SDR-GNN) for incomplete multimodal learning in conversational emotion recognition. SDR-GNN constructs an utterance semantic interaction graph using a sliding window based on both speaker and context relationships to model emotional dependencies. To capture higher-order and high-frequency information, SDR-GNN utilizes weighted relationship aggregation, ensuring consistent semantic feature extraction across utterances. Additionally, it performs multi-frequency aggregation in the spectral domain, enabling efficient recovery of incomplete modalities by extracting both high- and low-frequency information. Finally, multi-head attention is applied to fuse and optimize features for emotion recognition.
Extensive experiments on various real-world datasets demonstrate that our approach is effective in incomplete multimodal learning and outperforms current state-of-the-art methods.
\end{abstract}


\maketitle

\section{Introduction}

\begin{figure}[t]
	\centering
	\includegraphics[width=0.99\linewidth]{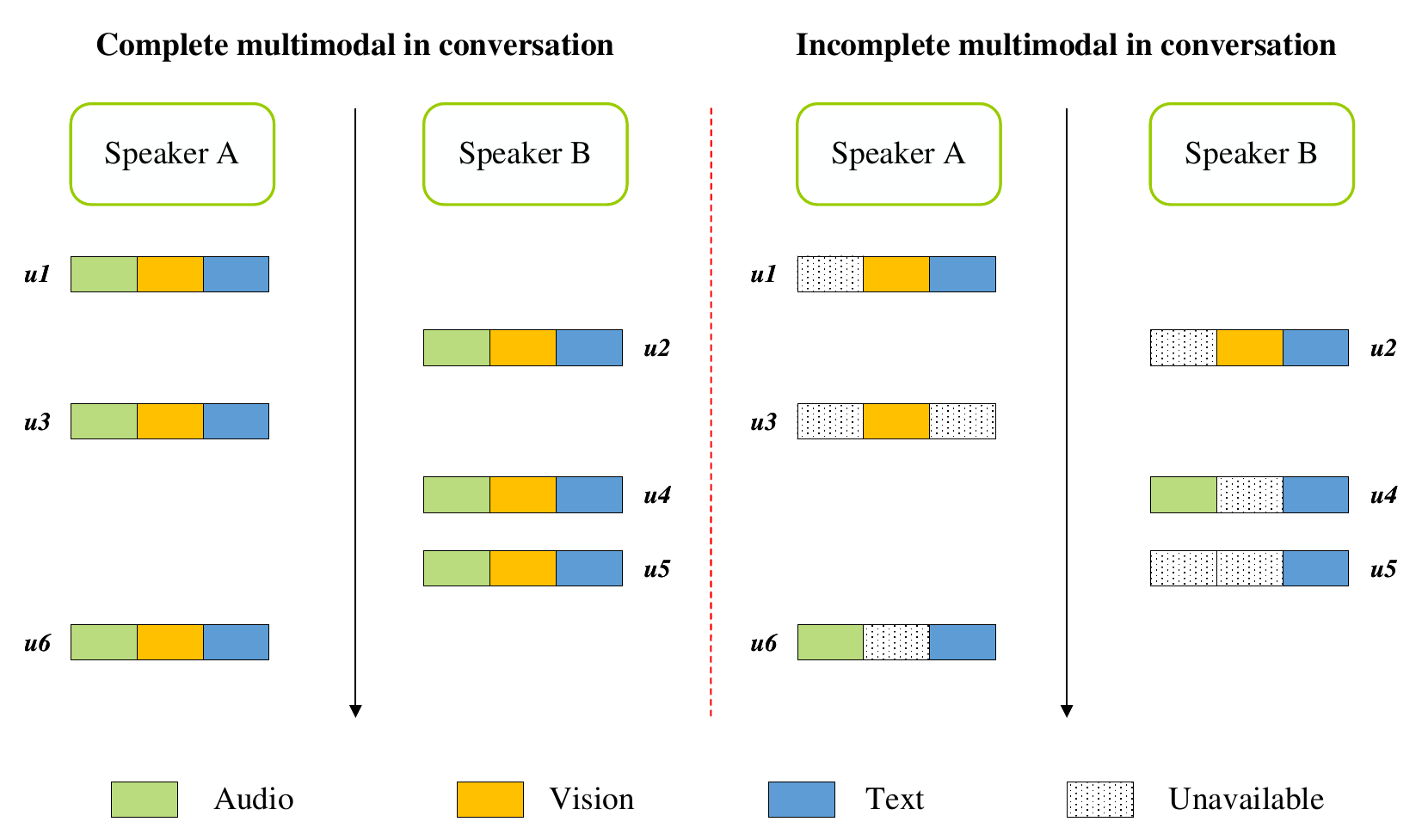}
	\caption{A toy example of complete multimodal features and incomplete multimodal features in conversation. Missing modalities pose a considerable challenge to capturing intra- and inter-modal semantic dependencies.}
	\label{Figure1}
\end{figure}

Multimodal Emotion Recognition in Conversations (ME\\RC) \cite{zhang2024multi, shou2022conversational, shou2025masked, shou2022object, shou2023comprehensive, shou2024adversarial} aims to identify the emotions expressed by each multimodal utterance in conversation scenes. Unlike traditional Unimodal Emotion Recognition in Conversations (UERC) \cite{nie2021gcn, meng2024deep, shou2023adversarial, ai2023gcn, ai2024gcn}, MERC can use textual, auditory, and visual modal information from the utterance to reveal more realistic emotions of the speaker by capturing the consistency and complementary semantics within and between modalities \cite{fan2024fusing,yang2024emotion,meng2024multi, shou2024contrastive, shou2024spegcl, shou2024efficient}. With the development of human-computer interaction, MERC has attracted significant attention from researchers because it can be widely used to understand and generate conversation \cite{chatterjee2019semeval, ying2021prediction, shou2023graph, shou2023czl}. However, most existing MERC methods usually assume that each utterance has complete modalities, ignoring the incomplete modality problem \cite{hu2021mmgcn, chen2023multivariate, meng2024masked, shou2024revisiting}. Unfortunately, obtaining complete multimodal data is incredibly challenging in practical conversation scenarios \cite{pham2019found}. For example, auditory data may not be available due to noise interference, visual data may not be available due to light or occlusion, and even more modal data may not be available due to sensor failure \cite{Wang_2023_ICCV, ai2024edge, zhang2024multi}. Fig. \ref{Figure1} presents a sample conversation between two speakers, where each utterance contains three modalities. The conversation on the right side illustrates the condition when modalities are incomplete.

The problem of incomplete modalities poses significant challenges for MERC tasks. To this end, researchers have proposed various methods to solve this problem mainly from how to perform modal recovery \cite{pham2019found, Wang_2023_ICCV, sun2023efficient, lian2023gcnet}. For instance, \textit{Pham et al.} \cite{pham2019found, ai2023two, meng2024revisiting, ai2024mcsff} proposed the MCTN model considering the semantic consistency between modalities. MCTN constructs cyclic transformations between modalities through sequence modeling to learn robust joint representations and uses cyclic consistency loss to achieve modality recovery. \textit{Wang et al.} \cite{Wang_2023_ICCV} considered the consistency of distribution between modalities and proposed the DiCMoR model. DiCMoR reduces the distribution gap by mapping different modalities to a latent space with Gaussian distribution and samples the characteristic distribution of the latent space to achieve modal recovery. \textit{Sun et al.} \cite{sun2023efficient, shou2023graphunet, ai2024graph, shou2024low, shou2024graph} considered the consistency of long-range semantics between modalities and proposed the EMT-DLFR model. EMT-DLFR captures consistent semantics in the global dialogue context by building a multi-modal Transformer and achieves modality recovery through feature reconstruction. \textit{Lian et al.} \cite{lian2023gcnet} considered the complex relationship between multi-modal utterances and proposed the GCNet model. GCNet uses graph neural networks to model context and speaker relationships separately to capture consistent semantics for missing modality recovery. Although these methods show good performance, they still have some limitations:

 \textbf{\textcolor{black}{(i) Limitations in capturing higher-order information.} }
In single-modal or multi-modal emotion recognition, the distribution of modalities in conversations is typically fixed. However, in conversations with incomplete modalities, the absence of modalities is often unpredictable. Models need to adapt to modalities absence of varying degrees and under different circumstances. \textcolor{black}{While existing graph-based models, including GCNet \cite{lian2023gcnet, ai2024seg, ai2024contrastive}, do capture higher-order information through information propagation, they rely on traditional graph structures, which are limited to binary relationships between nodes. These fixed structure graphs often struggle to capture complex semantic dependencies in conversations, especially when adapting to various missing modalities.} MMIN \cite{zhao2021missing} proposed six possible missing-modality conditions, but it can only learn for individual utterances. \textcolor{black}{In contrast, our approach utilizes a hypergraph structure, which effectively models higher-order relationships among multiple nodes \cite{bai2021hypergraph}. This allows the model to capture more complex and nuanced dependencies, overcoming the limitations of conventional graphs that GNN-based models face.} Therefore, how to capture the complex semantic dependencies between utterances, adapt to different situations, and optimize the recovery of incomplete modalities is an issue that cannot be ignored.

 \textbf{\textcolor{black}{(ii) Limitations in handling high-frequency information.}} 
Much research shows that high-frequency signals that reflect dissimilarity are as crucial as low-frequency signals that reflect consistency in MERC tasks \cite{hu2021mmgcn, chen2023multivariate}. Because the message propagation of GNN \cite{bo2021beyond} has low-pass filtering characteristics, node representation is achieved by aggregating consistent low-frequency information in the neighborhood and suppressing differential high-frequency information. \textcolor{black}{This inclination towards low-frequency components results in over-smoothing, where distinctive emotional transitions—the high-frequency signals—are suppressed, masking important intra-modal shifts.} Regrettably, the constructed utterances-emotion interaction graphs often have semantic inconsistencies, and it is crucial to retain high-frequency information. \textcolor{black}{Our proposed SDR-GNN addresses this by preserving and leveraging high-frequency information to capture rapid transitions and local changes, which are integral for comprehensive emotional analysis. Consequently,} simultaneously retaining and fusing high- and low-frequency information to guide the recovery of incomplete modalities is a challenge that must be overcome.

Inspired by the above analysis, this paper proposes a novel Spectral Domain Reconstruction Graph Neural Network for Incomplete Multimodal Learning in Conversational Emotion Recognition, named SDR-GNN.  SDR-GNN can capture the complex emotional dependencies between utterances while learning multi-frequency information in multimodal features for incomplete modal recovery to obtain better emotion recognition results. Specifically, SDR-GNN first simulates the modal missing problem in real conversation scenarios by randomly discarding some modal features, and adding speaker information to the discourse features to form multimodal nodes. Subsequently,  to model the complex semantic dependencies between multimodal utterances, SDR-GNN constructs the emotional interaction graph from the context and speaker relationships based on a sliding window, where the nodes in the sliding window are fully connected and construct the context and speaker hyperedges separately. Next,  to capture the complex emotional dependence between far and near utterances and learn multi-frequency information in multimodal features, SDR-GNN uses a neighborhood relationship awareness layer, a hyperedge relationship awareness layer, and a multi-frequency information awareness layer separately for information propagation. Finally, SDR-GNN reconstructs based on the learned features to guide the recovery of incomplete modalities and uses multi-head attention for feature fusion to achieve emotion recognition. We conducted experiments on three conversational datasets, verifying the effectiveness of our method. The experimental results demonstrate that our SDR-GNN outperforms existing approaches. The main contributions of this paper can be summarized as follows:

\begin{itemize}
	\item \textcolor{black}{Existing graph neural networks (GNNs) are constrained by their inherent limitations, which may lead to over-smoothing and the erasure of high-frequency signals, making it difficult to fully utilize multi-frequency information. We have not only addressed this limitation but also applied our approach to multimodal emotion recognition under incomplete modalities, thereby filling the gap in current works.}
	
	\item We propose a novel framework, SDR-GNN, to deal with incomplete conversational data in the MERC task, which jointly considers the higher-order information of modalities and multi-frequency  features, and fully utilizes the semantic dependence in both speaker and context for missing modality recovery and emotion recognition.
	
	\item Experimental results on three benchmark datasets verify the effectiveness of our method. SDR-GNN outperforms existing state-of-the-art approaches in the domain of incomplete multimodal learning in conversational emotion recognition.
\end{itemize}

\section{Related Works}
\subsection{Multimodal Emotion Recognition in Conversations}

\textcolor{black}{Multimodal Emotion Recognition in Conversations has gained significant attention in recent years due to its potential applications in various fields. Multimodal ERC leverages multiple data modalities, including text, audio, and visual data, to capture and analyze emotions more comprehensively during conversational exchanges.}

\textcolor{black}{To better utilize multimodal information to address the ERC problem, researchers have proposed various methods. MulT \cite{tsai2019multimodal} model uses cross-modal transformers to capture long-range dependencies. MMGCN \cite{hu2021mmgcn} constructs a comprehensive graph to handle multimodal and extensive contextual information, and includes speaker embeddings to encode speaker-specific details. M2FNet \cite{chudasama2022m2fnet}, a multimodal network based on multi-head attention layers to capture crossmodal interactions. MultiEMO \cite{shi2023multiemo} model incorporates bidirectional multi-head cross-attention layers for effective fusion. What's more, CBERL \cite{meng2024deep} using a multimodal generative adversarial network to address the imbalanced distribution of emotion categories in raw data.}

\textcolor{black}{One major assumption in MERC is that data from all modalities are complete and continuous. However, in the real world, data from modalities are often incomplete due to various reasons, making learning under incomplete modalities a promising area of research.}

\subsection{Incomplete Multimodal Learning}
Multimodal learning aims to utilize information from a variety of data modalities to improve generalization performance. However, in some conditions, modalities may be missing or unavailable. A straightforward method is to use existing data for classification. Additionally, there are strategies that conduct data imputation aiming to reconstruct missing data. We divide the existing methods into two categories: non-reconstruction and reconstruction methods.

Existing non-reconstruction approaches primarily focus on the analysis of incomplete data, such as through maximizing correlations \cite{andrew2013deep,ma2021efficient,lin2021completer}. Hotelling et al. \cite{hotelling1992relations} introduced CCA, which maximizes canonical correlations by linearly mapping multimodal features into a low-dimensional space. In contrast to CCA's linear focus, Andrew et al. \cite{andrew2013deep} developed DCCA, which enhances traditional CCA by addressing its limitations related to linear associations. It employs deep neural networks to uncover more intricate, non-linear relationships across different modalities. Additionally, Wang et al. \cite{wang2015deep} introduced DCCAE. DCCAE advances CCA by incorporating autoencoders, which are designed to extract latent features from each modality. This approach optimizes both the reconstruction accuracy of the autoencoders and the canonical correlations, effectively balancing the integrity of modality-specific structures with the connectivity between modalities.

Reconstruction methods, on the other hand, aim to ensure data completeness, primarily through data imputation \cite{parthasarathy2020training, ma2021maximum, zhang2022deep}, generating missing data \cite{vincent2008extracting,tran2017missing, cai2018deep}, or reconstructing incomplete data by learning feature representations. Parthasarathy et al. \cite{parthasarathy2020training} proposed an attention-based model that fills missing video data with zero vectors. Zhang et al. \cite{zhang2022deep} developed CPM-Net, which integrates an encoderless model with a clustering-like classification loss to learn features and pads missing modalities with average values. Moreover, several DNN-based models have been developed, including autoencoders \cite{tran2017missing}, GAN \cite{wang2018partial}, and Transformers \cite{yuan2021transformer}.

To better reconstruct incomplete data, researchers started to explore feature representations. For example, Lian et al. proposed GCNet \cite{lian2023gcnet}, which utilizes GNN-based models to capture different types of information in conversations to reconstruct missing modalities. Wang et al. \cite{Wang_2023_ICCV} considered the consistency of data distributions to recover missing features.

\section{Methodology}
\begin{figure*}[t]
	\centering
	\includegraphics[width=0.98\linewidth]{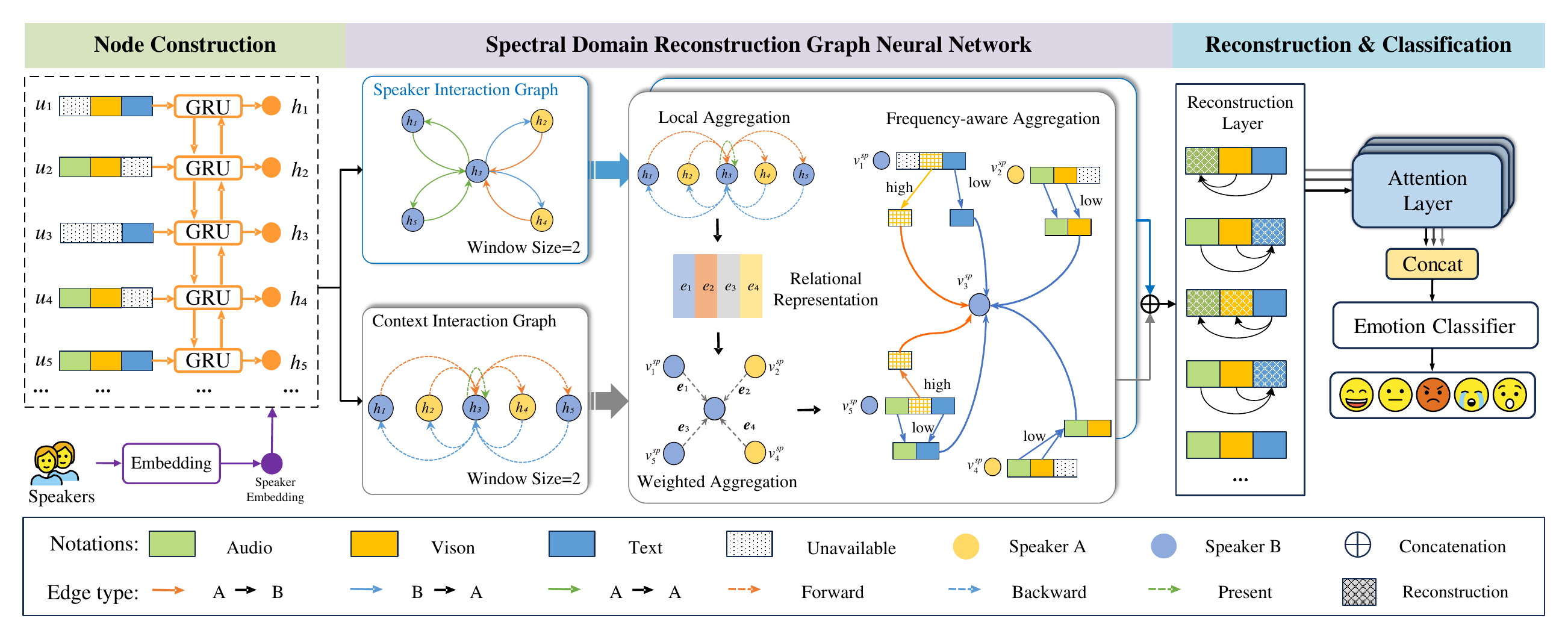}
	\caption{The overall structure of the framework. First, we encode features of the utterance using a Bi-GRU to obtain the contextual embedding of each node. Then, we apply the SDR-GNN to capture features, jointly considering higher-order and multi-frequency information. Finally, we reconstruct the incomplete features and classify the emotion labels.}
	\label{Figure2}
\end{figure*}

The main objective of MERC is to assign an appropriate emotion label to each utterance within a dialogue. This paper specifically addresses scenarios where multimodal data is incomplete—common in real-world applications where some modalities might be unavailable or lost due to technical issues. We introduce a novel framework, SDR-GNN, designed to effectively manage and process these incomplete datasets. Our approach leverages the intrinsic structure of conversational data and employs graph neural networks to interpolate or reconstruct the missing modalities, ensuring robust emotion recognition even with partial information. Fig. \ref{Figure2} in the paper provides a visual overview of the SDR-GNN framework, illustrating its key components and operational flow in handling missing multimodal features across conversational utterances. 

\subsection{Node Construction}
We define each conversation consists of a series of utterances $C = \{u_{1},u_{2},\cdots,u_{n}\}$, where $n$ is the number of utterances. Each conversation involves $N$ speakers $P = \{p_{1},p_{2},\cdots,p_{N}\}$($N\geq2$). Each utterance $u_{i}$ is spoken by $p_{s(u_i)}$, where the function $s(\cdot)$ maps the index of utterance into its corresponding speaker. For each utterance $u_{i}$, we extract multimodal features $u_{i}=\{\eta{f}_{i}^m\}_{m\in\{a, v, t\}}$. Here, $f_{i}^a \in \mathbb{R}^{d_a}$, $f_{i}^v \in \mathbb{R}^{d_v}$ and $f_{i}^t \in \mathbb{R}^{d_t}$ represent the audio, visual and text features of the utterance, respectively. $\{d_{m}\}_{m\in\{a, v, t\}}$ is the feature dimension of each modality. Each $\eta$ of $u_{i}$ is defined as follows:	

\begin{equation}
	\label{eq1}
	\eta=\begin{cases}
		1, {f}_{i}^m\text{ is available;} \\
		0, \text{otherwise.}\\
	\end{cases}
\end{equation}	

In this paper, we assume at least one modality-complete data is available for analysis. Therefore, an incomplete $M$-modal dataset has $\left(2^{M}-1\right)$ different missing patterns, in line with previous works \cite{zhang2022deep,lian2023gcnet}.

We employ a bidirectional Gated Recurrent Unit (GRU) to extract contextual features and dynamically analyze dependency relationships. The computation is performed as follows:

\begin{equation}
	\begin{aligned}
		&u_{i} = BiGRU(u_{i},h_{i(+,-)1}),\\
		&H = \{h_i\}_{i=1}^n \in \mathbb{R}^{n \times (d_a + d_v + d_t)},
	\end{aligned}
\end{equation}

$H$ is the matrix containing all hidden states $h_i$ for $(i=1)$ to $n$. Each hidden state is a vector that captures contextual information up to the $i$-th position from both directions of the sequence.

\subsection{Spectral Domain Reconstruction Graph Neural Network}
\label{sec3-2}
The main idea of Spectral Domain Reconstruction Graph Neural Network is to capture the multivariate relationships between domain nodes, resulting in better aggregation effects for the following reconstruction task. We first construct relation graph convolutional networks (R-GCN) \cite{schlichtkrull2018modeling} in capturing node features, capturing both contextual and speaker features. In addition, considering the dynamic absence of modalities, we construct a hypergraph with edge-dependent node weights to flexibly aggregate node information.
Recent works has verified the effectiveness of multi-frequency emotional information in the ERC task \cite{chen2023multivariate,bo2021beyond}, therefore we design a frequency-aware module specifically to capture this information.

We have developed speaker interaction graphs and context interaction graphs as the primary modules for extracting emotion cues. In these graphs, edges measure the significance of connections between nodes, where the type of edge determines the propagation method of various information. While both the speaker and context graphs use identical edges, each edge represents a distinct dependency.

\textbf{Edges:} Considering the overwhelming number of connections when each node interacts with all others, we streamline this by limiting node interactions to a fixed-size context window $w$, following insights from previous research that emphasize the importance of local context. Therefore, a node $v_i$ only connects with nearby nodes within the context window $\{v_j\}_{j \in \left[max(i-w, 1), min(i+w, L)\right]}$, significantly reducing complexity. We select $w$ from the set $\{1, 2, 3, 4\}$ and denote the edge from node $v_i$ to $v_j$ as $e_{ij}\in\mathcal{E}$ ($|\mathcal{E}|=n+2w-1$).

\textbf{Speaker Interaction Graph:} The speaker interaction graph leverages the various speakers and their corresponding utterances to map out the dependencies among speakers within a conversation. Each edge $e_{ij}$ in the graph is tagged with a speaker identifier $\alpha_{ij}$ from the set \text{$\alpha$}, which encompasses all speaker types present in the dialogue. The cardinality of \text{$\alpha$}, represented as \text{$\alpha$}, indicates the number of distinct speaker types. For each connection $e_{ij}$, $\alpha_{ij}$ denotes the directional flow from speaker $p_{s(u_i)}$ to speaker $p_{s(u_j)}$, where $p_{s(u_i)}$ and $p_{s(u_j)}$ are the speaker identifiers for $u_i$ and $u_j$, respectively.

\textbf{Context Interaction Graph:} Context interaction graph utilizes contextual information to delineate the contextual dependencies within a conversation. 
Each edge $e_{ij}$ is assigned a context type identifier $\beta_{ij} \in$ \text{$|\beta|$}, which contains all possible context types in the discussion. The determination of \text{$\beta$} values is influenced by the relative positioning of $u_i$ and $u_j$ within the dialogue, with possible values including {backward, present, forward}. Therefore, the total number of context types, \text{$|\beta|$}, is three.

\textbf{Weighted HyperGraph:} In hypergraphs, we define two types of weights: an edge weight, $\lambda(e)$, for each edge $e$, and a node weight, $\gamma_e(v)$, for each node $v$ incident to edge $e$, also known as edge-dependent node weight. Intuitively, $\gamma_e(v)$ represents the contribution of node $v$ to the hyperedge $e$, enriching the representation of detailed multimodal and contextual dependencies. Consequently, edge-dependent node weights are expressed using a weighted incidence matrix. $\hat{\mathbf{H}}\in\mathbb{R}^{n\times|\mathcal{E}|}$:

\begin{equation}
	\hat{\mathbf{H}}=\begin{cases}\gamma_e(v)\text{, edge }\text{is incident with node}~v;\\\quad0\text{, otherwise}.\end{cases}
\end{equation}

\textbf{Graph learning:} We use R-GCN to aggregate the local information in the graph, then use hypergraph for weighted aggregation. The calculation is shown as follows:

\begin{equation}
	v_i^{sp}=ReLU\left(\sum_{r\in\alpha}\sum_{j \in {N}_i^r}{\frac{1}{\left|N_i^r\right|}}W_1^rh_j\right),
\end{equation}

\begin{equation}
	v_i^{co}=ReLU\left(\sum_{r\in\beta}\sum_{j \in {N}_i^r}{\frac{1}{\left|N_i^r\right|}}W_2^rh_j\right),
\end{equation}

\begin{equation}
	\mathbf{V}^{(l+1)}=LeakyReLU(\mathbf{D}^{-1}\mathbf{H}\mathbf{W}_{e}\mathbf{B}^{-1}\mathbf{\hat{H}}\mathbf{V}^{(l)}),
\end{equation}

where $v_i^{sp}\in\mathbb{R}^{h}$, $v_i^{co}\in\mathbb{R}^{h}$ denote the outputs of nodes in Speaker types and Context types, respectively. $N_i^r$ denotes the set of all neighbor nodes of $v_i$ under relation $r$, and $|N_i^r|$ is the number of $N_i^r$. $W_1^r$ and $W_2^r$ are the trainable parameters for different types of graph under relation $r$, respectively.
$\mathbf{H}\in\mathbb{R}^{n\times|\mathcal{E}|}$  represent the incidence matrix, in which a nonzero entry $\mathbf{H}_{ve}$ = 1 indicates that the edge $e$ is incident with the node $v$; otherwise $\mathbf{H}_{ve}$ = 0.
$\mathbf{D}_{\mathcal{H}}\in\mathbb{R}^{n\times n}$ and $\mathbf{B}\in\mathbb{R}^{|\mathcal{E}|\times|\mathcal{E}|}$ are the node degree matrix and edge degree matrix, respectively.
$\mathbf{V}^{(l)}\:=\:\{v_{i,(l)}\}_{i=1}^n\in\mathbb{R}^{n\times (d_a + d_v + d_t)}$ is the input at layer $l$.
$\mathbf{W}_e= $diag$(\lambda(e_1),\cdots,\lambda( e_{ij}) )$ is the edge weight matrix.

\textbf{Frequency-Aware Graph:}
Although speaker graph and context graph can capture feature dependencies, they still follow the generic graph learning protocol, which aggregates and smooths signals from the local neighborhood, thereby erasing high-frequency signals \cite{chen2023multivariate,bo2021beyond}. These signals can be crucial for ERC tasks. To effectively learn different types of frequency information between the central node and its neighbors, we designed a self-gating mechanism. Specifically, it calculates the correlation between the central node and its neighbors, learning the multi-frequency information of multimodal features.
Mathematically:

\begin{equation}
	\overline{v}_{ij}^x = Concat(v_i^x,v_j^x), x\in\{sp,co\}
\end{equation}

\begin{equation}	l_{i}^{sp}=v_{i}^{sp}+\sum_{r\in\alpha}\sum_{j\in{N}_{i}^r}\tanh\left(\frac{W_{3}^r\overline{v}_{ij}^{sp}}{\sqrt{|N_i^r||N_j^r|}}\right)v_{j}^{sp},
\end{equation}

\begin{equation}
	l_{i}^{co}=v_{i}^{co}+\sum_{r\in\beta}\sum_{j\in{N}_{i}^r}\tanh\left(\frac{W_{4}^r\overline{v}_{ij}^{co}}{\sqrt{|N_i^r||N_j^r|}}\right)v_{j}^{co},
\end{equation}

Here, $W_{3}^r, W_{4}^r \in \mathbb{R}^{2h}$ are trainable weight matrices, and $\tanh(\cdot)$ is the hyperbolic tangent function, which scales the input to the range [-1,1]. \textcolor{black}{In the context of graph neural networks, low-frequency signals can be thought of as generalized information propagated across large areas of the network, indicating similarity or commonality among nodes. High-frequency signals, conversely, emphasize differences or specific characteristics distinct to neighboring nodes. These signals are derived through the spectral decomposition of the graph Laplacian, which allows us to separate these frequency components mathematically.} Through this mechanism, the outputs of $W_{3}^r\overline{v}_{ij}^{sp}$ and $W_{4}^r\overline{v}_{ij}^{co}$ effectively gauge the significance of various frequency components. \textcolor{black}{The self-gating mechanism, as proposed in our SDR-GNN, enables dynamic differentiation and integration of these frequency signals, helping retain essential low-frequency information while preserving critical high-frequency details crucial for tasks involving nuanced data.} For example, if $W_{3}^r\overline{v}_{ij}^{sp} < 0$, high-frequency messages are prominent, signifying a greater difference between node $i$ and its neighbor $j$, and vice versa.

To aggregate these representations, we concatenate them to form the final representation of the Local Enhancement Graph:

\begin{equation}
	\overline{l_i} = Concat(l_{i}^{sp},l_{i}^{co}).
\end{equation}

\subsection{Reconstruction \& Self Optimization}
To better utilize multi-frequency data, we input the extracted features into a linear transformation layer for predicting missing data and achieving data recovery. Then input the recovered data into a multi-head attention layer \cite{vaswani2017attention} to fuse and optimize reconstructed modalities, which can be shown as follows:

\begin{equation}
	\hat{F} = \overline{L}W_m+b_m, m\in\{a,v,t\},
\end{equation}

\begin{equation}
	\overline{F_i} = \mathrm{softmax}(\frac{QK^{T}}{\sqrt{d}})V,
\end{equation}

\noindent where ${\overline{L}}=\{\overline{l}_i\}_{i=1}^n\in\mathbb{R}^{n\times d_h}$ is the matrix containing all hidden states $\overline{l}_i$ and $\hat{F}^m=\{\hat{f}_i\}_{i=1}^n\in\mathbb{R}^{L\times d_m}$ is the estimated complete data.
$W_m\in \mathbb{R} ^{d\times d_m}$ and $b_m\in \mathbb{R} ^{d_m}$ are the trainable parameters, where $d_m$ is the feature dimension for each modality.
For the attention layers, $Q=\hat{F}W_Q$, $K=\hat{F}W_K$, $V=\hat{F}W_V$. $Q=\hat{F}W_Q$, $K=\hat{F}W_K$, $V=\hat{F}W_V$ are the trainable parameter matrices. In this approach, multiple attentions are combined to obtain the output results of the multi-head attention layer as follows:

\begin{equation}
	Multihead(\overline{F})=Concat(\overline{F_1},\ldots,\overline{F_k})W,
\end{equation}

\noindent where $\overline{F_1},\ldots,\overline{F_k}$ is the output of each attention layer, $k$ is the number of attention layers, and $W$ is the trainable parameter matrix.

\subsection{Emotion Classifier}
To enhance the learning of more discriminative features for conversation understanding, we input the latent representations $\overline{L}=$ $\{\overline{l}_i\}_{i=1}^n$ into a fully-connected layer, subsequently followed by a softmax layer to compute the classification probabilities:

\begin{equation}
	\hat{Y}=softmax(\overline{L}W_{c}+b_{c}),
\end{equation}

\noindent here $\hat{Y}=\{\hat{y}_i\}_{i=1}^n\in\mathbb{R}^{n\times c}$ is the estimated probabilities, $y_i \in \{1,\ldots,c\},\hat{y}_i\in \{1,\ldots,c\}$. Where $y_i$ is the true labels and $c$ is the number of discrete labels in the corpus, $W_{c}\in \mathbb{R} ^{d\times c}$ and $b_{c}\in \mathbb{R} ^{c}$ are the trainable parameters.
$W_{c}\in \mathbb{R} ^{d\times c}$ and $b_{c}\in \mathbb{R} ^{c}$ are the trainable parameters.

Our loss function consists of two parts, the reconstruction function and the cross entropy function. The reconstruction function is used to calculate the difference between the original data and the filled data, while the cross entropy function is used for label classification. The calculation is illustrated as follows:
\begin{equation}
	\mathcal{L}_{ce}=-\frac{1}{n}\sum_{i=1}^{n}y_{i}\log(\hat{y}_{i}),
\end{equation}

\begin{equation}
	\mathcal{L}_{rec}=\sum_{\textcolor{black}{m\in\{a,v,t\}}}\frac{1}{d_mn}\sum_{i=1}^{n}\|(\hat{f}_i^m-f_i^m)\|^2,
\end{equation}

\begin{equation}
	\mathcal{L}=\textcolor{black}{(1-e)}\mathcal{L}_{ce}+\textcolor{black}{e}\mathcal{L}_{rec}.
\end{equation}

\section{Experiments}
In this section, we describe the three benchmark conversational datasets employed in our experiments, explain the evaluation metrics and multimodal features used, and introduce a variety of advanced baselines for comparison in the context of incomplete multimodal learning.

\subsection{Datasets}
To assess the efficacy of the SDR-GNN model, we conducted experiments using three benchmark conversational datasets: IEMOCAP \cite{busso2008iemocap}, CMU-MOSI \cite{zadeh2016multimodal}, and CMU-MOSEI \cite{zadeh2018multimodal}. These datasets are widely recognized in the research community for their comprehensive coverage of emotional and multimodal human interactions, making them ideal for testing new models in multimodal learning contexts.

\textbf{IEMOCAP} includes multiple conversations between two speakers, segmented into short utterances each annotated with discrete emotion labels. For consistency in comparisons, we employ two prevalent labeling methods, generating datasets with either four or six classes. The four-class dataset includes the emotions: anger, happiness (where excitement is merged with happiness), sadness, and neutral \cite{poria2017context}. The six-class dataset encompasses: anger, happiness, sadness, neutral, excitement, and frustration \cite{majumder2019dialoguernn}.

\textbf{CMU-MOSI} features a collection of movie review videos from online platforms, comprising 2,199 short monologue clips. Each clip is rated with a sentiment intensity score on a scale from -3 (most negative) to +3 (most positive).

\textbf{CMU-MOSEI} extends CMU-MOSI by incorporating a wider range of topics with 22,856 movie review clips from YouTube, maintaining the same sentiment scoring method from -3 to +3.

\begin{table}[h]
	\centering
	\renewcommand\tabcolsep{3.2pt}
	\renewcommand\arraystretch{1.20}
	\caption{Statistical information on IEMOCAP, CMU-MOSI and CMU-MOSEI.}
	\label{Table1}
	\begin{tabular}{cc|c|c|c|c}
		\hline
		\multicolumn{2}{c|}{\multirow{2}{*}{Dataset}} & \multicolumn{2}{c|}{\# utterances} & \multicolumn{2}{c}{\# conversations} \\
		&& train \& val		& test	& train \& val 	& test 	\\ \hline \hline
				
		\multicolumn{2}{c|}{IEMOCAP(four-class)} & 4290  &1241 & 120  & 31  	 \\
		\hline
				
		\multicolumn{2}{c|}{IEMOCAP(six-class)}  & 5810  &1623 & 120  & 31    \\
		\hline
		
		\multicolumn{2}{c|}{CMU-MOSI}     & 1513  & 686 & 62  & 31  	\\
		\hline
		
		\multicolumn{2}{c|}{CMU-MOSEI} 	 & 18197  & 4659   & 2549 & 676    \\
		\hline
	\end{tabular}
\end{table}

\subsection{Implementation Details and Evaluation Metrics}

We evaluate the performance of various methods on multimodal datasets with different missing rates, defined as $\mathcal{M} = 1-\frac{\sum_{i=1}^{n}{s_i}}{n \times M}$. Here, $s_i$ represents the number of available modalities for the $i^{th}$ sample, $L$ is the total number of samples, and $M$ is the total number of modalities. For each sample, modalities are randomly masked according to $\mathcal{M}$, ensuring at least one modality per sample. This constraint results in $\mathcal{M} \leqslant \frac{M-1}{M}$. For $M=3$, $\mathcal{M}$ ranges from $0.0$ to $0.7$, the latter approximating $\frac{M-1}{M}$. In line with prior research \cite{Wang_2023_ICCV,lian2023gcnet}, the missing rate remains constant across training, validation, and testing phases.

We utilize the datasets IEMOCAP, CMU-MOSI, and CMU-MOSEI, which are equipped with predefined splits for training, validation, and testing. The model configuration that performs optimally is identified using the validation set and subsequently evaluated on the test set. Our methodology involves adjusting two key parameters: the dimension of latent representations, labeled as \(h\), and the size of the interaction window, labeled as \(w\). Our experiments involve values of \(h \in \{100, 150, 200, 250\}\) and \(w \in \{1, 2, 3, 4\}\), applied across all datasets. For optimization, the Adam optimizer is employed, with a learning rate of 0.001 and a weight decay of 0.00001.
 Additionally, we incorporate a multi-head attention mechanism with $k = 256$ heads. To mitigate overfitting, Dropout \cite{srivastava2014dropout} is applied at a rate of $p = 0.5$. The reliability of our results is ensured by averaging the performance over ten trials on the test set.

To verify our method, we select the following evaluation metrics to fair compete with different approaches.

For \textbf{IEMOCAP}, we choose weighted average F1-score (WAF1) as the evaluation metric. \textcolor{black}{WAF1 is calculated as a weighted mean F1 over different emotion categories with weights proportional to the number of utterances in each emotion class, which can be shown as follows}, in line with previous works \cite{majumder2019dialoguernn,lian2023gcnet}.

\begin{equation}
	\color{black}
	WAF1=\frac{\sum_{j=1}^EN_j*F1_j}{\sum_{j=1}^EN_j}
\end{equation}

\textcolor{black}{where $E$ is the total number of emotion categories, $N_j$ is the number of samples in category $j$, and $F1_j$ is the F1 score for category $j$. }

For \textbf{CMU-MOSI} and \textbf{CMU-MOSEI}, we focus on the negative/positive classification task, with scores assigned to less than 0 for negative and greater than 0 for positive, respectively. We choose WAF1 as the primary metric and the accuracy (ACC) of the classification task as the secondary metric.

\subsection{Baselines}
\noindent\textbf{CCA}\cite{hotelling1992relations}: CCA aims to find the linear relationships with the maximum correlation between two multimodal datasets. By linearly mapping them into a low-dimensional common space, CCA learns the relationships between different modalities. It is a strong benchmark model, especially suitable for scenarios where linear relationships can capture the interactions between modalities well.

\noindent\textbf{DCCA}\cite{andrew2013deep}: DCCA enhances traditional CCA by addressing its limitations related to linear associations. It employs deep neural networks to uncover more intricate, non-linear relationships across different modalities.

\noindent\textbf{DCCAE}\cite{wang2015deep}: DCCAE advances CCA by incorporating autoencoders, which are designed to extract latent features from each modality. This approach optimizes both the reconstruction accuracy of the autoencoders and the canonical correlations, effectively balancing the integrity of modality-specific structures with the connectivity between modalities.

\noindent\textbf{AE}\cite{bengio2007greedy}: In incomplete multimodal learning, autoencoders are widely used to impute missing data from partially observed inputs. By jointly optimizing the reconstruction loss of autoencoders and the classification loss of downstream tasks, this method supports a trade-off in implementation.

\noindent\textbf{CRA}\cite{tran2017missing}: CRA extends AE by integrating a series of residual autoencoders into a cascaded architecture for data imputation. During implementation, CRA optimizes both imputation and downstream tasks in an end-to-end manner, enhancing the quality of data completion and the performance of tasks.

\noindent\textbf{MMIN}\cite{zhao2021missing}: The MMIN model integrates CRA with cycle consistency learning to predict the latent representations of missing modalities. This approach makes MMIN a robust benchmark model, demonstrating excellent performance under a range of missing conditions. This dual-component strategy enhances the model's ability to handle incomplete data, ensuring more accurate and reliable predictions across different scenarios.

\noindent\textbf{CPM-Net}\cite{zhang2022deep}: CPM-Net accounts for both completeness and versatility in multi-view representation to learn discriminative latent features. The framework is constructed to optimize the use of multiple partial views by defining and theoretically proving "completeness" and "versatility" in multi-view representations.

\noindent\textbf{MCTN}\cite{pham2019found}: MCTN is a method designed to learn robust joint representations by translating between modalities. It combines an autoencoder with a cycle consistency loss to achieve modality reconstruction.

\noindent\textbf{GCNet}\cite{lian2023gcnet}: GCNet is a state-of-the-art method that utilizes graph neural networks to capture different types of features and recover missing modalities, further improving the performance of downstream tasks.

\noindent\textbf{DiCMoR}\cite{Wang_2023_ICCV}: DiCMoR is also a state-of-the-art method which considers the consistency of data distributions to recover the missing features, in order to obtain better recovered data.

\begin{table*}[b]
	\centering
	\renewcommand\tabcolsep{6.5pt}
	\renewcommand\arraystretch{1.20}
	\caption{Comparison of performance with various missing rates on IEMOCAP. We report WAF1 scores(\%). Higher WAF1 indicates better performance. The best performance is highlighted in bold.}
	\label{Table2}
	\begin{tabular}{ccccccccccccc}
		\hline
		\multirow{2}{*}{Dataset}													&
		\multirow{2}{*}{Method}														&
		\multicolumn{9}{c}{\begin{tabular}[c]{@{}c@{}}Missing Rate\end{tabular}}	\\ \cline{3-11}
		& & 0.0	&0.1 &0.2 &0.3 &0.4 &0.5 & 0.6 & 0.7 & Average \\
		\hline \hline

		\multirow{8}{*}{\begin{tabular}[c]{@{}c@{}}IEMOCAP \\ (four-class)\end{tabular}}	
		&CCA$^\dag$    \cite{hotelling1992relations} & 64.52 & 65.19 & 62.60 & 59.35 & 55.25 & 51.38 & 45.73 & 30.61 & 54.33 \\
		&DCCA$^\dag$   \cite{andrew2013deep}       	 & 60.03 & 57.25 & 51.74 & 42.53 & 36.54 & 34.82 & 33.65 & 41.09 & 44.71 \\
		&DCCAE$^\dag$  \cite{wang2015deep}        	 & 63.42 & 61.66 & 57.67 & 54.95 & 51.08 & 45.71 & 39.07 & 41.42 & 51.87 \\
		&CPM-Net$^\dag$\cite{zhang2022deep}          & 58.00 & 55.29 & 53.65 & 52.52 & 51.01 & 49.09 & 47.38 & 44.76 & 51.46 \\
		&AE$^\dag$     \cite{bengio2007greedy}       & 74.82 & 71.36 & 67.40 & 62.02 & 57.24 & 50.56 & 43.04 & 39.86 & 58.29 \\
		&CRA$^\dag$    \cite{tran2017missing}      	 & 76.26 & 71.28 & 67.34 & 62.24 & 57.04 & 49.86 & 43.22 & 38.56 & 58.23 \\
		&MMIN$^\dag$   \cite{zhao2021missing}        & 74.94 & 71.84 & 69.36 & 66.34 & 63.30 & 60.54 & 57.52 & 55.44 & 64.91 \\
		&GCNet$^\dag$  \cite{lian2023gcnet}          & 78.36 & 77.48 & 77.34 & 76.22 & 75.14 & 73.80 & 71.88 & 71.38 & 75.20 \\
		
		&\textbf{Ours}  & \textbf{79.58} &\textbf{78.55} &\textbf{78.08} &\textbf{77.53} &\textbf{77.09} &\textbf{75.84} &\textbf{75.03} &\textbf{74.41} &\textbf{77.01}\\
		\hline
		\multirow{8}{*}{\begin{tabular}[c]{@{}c@{}}IEMOCAP \\ (six-class)\end{tabular}}	
		&CCA$^\dag$    \cite{hotelling1992relations} & 43.04 & 46.06 & 43.86 & 41.66 & 37.13 & 34.94 & 32.06 & 21.80 & 37.57 \\
		&DCCA$^\dag$   \cite{andrew2013deep}         & 42.18 & 39.15 & 34.47 & 27.65 & 23.69 & 22.86 & 22.71 & 27.38 & 30.01 \\
		&DCCAE$^\dag$  \cite{wang2015deep}	         & 46.19 & 43.77 & 41.28 & 37.98 & 34.58 & 30.02 & 26.78 & 27.66 & 36.03 \\
		&CPM-Net$^\dag$\cite{zhang2022deep}	         & 41.05 & 37.33 & 36.22 & 35.73 & 35.11 & 33.64 & 32.26 & 31.25 & 35.32 \\
		&AE$^\dag$     \cite{bengio2007greedy}	     & 56.76 & 52.82 & 48.66 & 42.26 & 35.18 & 29.12 & 25.08 & 23.18 & 39.13 \\
		&CRA$^\dag$    \cite{tran2017missing}	     & 58.68 & 53.50 & 49.76 & 45.88 & 39.94 & 32.88 & 28.08 & 26.16 & 41.86 \\
		&MMIN$^\dag$   \cite{zhao2021missing} 	     & 56.96 & 53.94 & 51.46 & 48.42 & 45.60 & 42.82 & 40.18 & 37.84 & 47.15 \\
		&GCNet$^\dag$  \cite{lian2023gcnet}	         & 58.64 & 58.50 & 57.64 & 57.08 & 56.12 & 54.40 & 53.60 & 53.46 & 56.18 \\
		
		&\textbf{Ours} &\textbf{61.34}  &\textbf{60.86} &\textbf{59.83} &\textbf{59.49}  &\textbf{59.16} &\textbf{57.38} &\textbf{55.51} &\textbf{55.26} &\textbf{58.60}\\
		\hline
		\hline
		\text{$^\dag$results come from \cite{lian2023gcnet}}
	\end{tabular}
\end{table*}

\begin{table*}[h]
	\centering
	\renewcommand\tabcolsep{0.25pt}
	\renewcommand\arraystretch{1.4}
	\caption{Comparison of performance with various missing rates on CMU-MOSI and CMU-MOSEI. We report WAF1/ACC scores(\%). The best performance is highlighted in bold.}
	\label{Table3}
	\begin{tabular}{ccccccccccccc}
		\hline
		\multirow{2}{*}{Dataset}													&
		\multirow{2}{*}{Method}														&
		\multicolumn{9}{c}{\begin{tabular}[c]{@{}c@{}}Missing Rate\end{tabular}}	\\ \cline{3-11}
		& & 0.0	&0.1 &0.2 &0.3 &0.4 &0.5 & 0.6 & 0.7 & Average \\
		\hline
		\hline

		\multirow{6}{*}{\begin{tabular}[c]{@{}c@{}}CMU-MOSI\end{tabular}}	
		
		&DCCA$^\star$  \cite{andrew2013deep}  & 75.4/75.3 & 72.2/72.1 & 69.1/69.3 & 65.2/65.4 & 62.0/62.8 & 59.9/60.9 & 57.3/58.6 & 56.0/57.4 & 64.6/65.2 \\
		&DCCAE$^\star$ \cite{wang2015deep}    & 77.4/77.3 & 74.7/74.5 & 71.9/71.8 & 66.7/67.0 & 62.8/63.6 & 61.3/62.0 & 58.8/59.6 & 57.4/58.1 & 66.3/66.7 \\
		&MCTN$^\star$  \cite{pham2019found}   & 81.5/81.4 & 78.5/78.4 & 75.7/75.6 & 71.2/71.3 & 67.6/68.0 & 64.8/65.4 & 62.5/63.8 & 59.0/61.2 & 70.1/66.7 \\
		&MMIN$^\star$  \cite{zhao2021missing} & 84.4/84.6 & 81.8/81.8 & 79.1/79.0 & 76.2/76.1 & 71.6/71.7 & 66.5/67.2 & 64.0/64.9 & 61.0/62.8 & 73.1/73.5 \\
		&GCNet$^\star$ \cite{lian2023gcnet}   & 85.1/85.2 & 82.3/82.3 & 79.5/79.4 & 77.2/77.2 & 74.4/74.3 & 69.8/70.0 & 66.7/67.7 & 65.4/65.7 & 75.1/75.2 \\
		&DiCMoR$^\star$\cite{Wang_2023_ICCV}  & 85.6/85.7 & 83.9/83.9 & \textbf{82.0}/\textbf{82.1} & 80.2/80.4 & 77.7/77.9 & \textbf{76.4}/\textbf{76.7} & \textbf{73.0}/\textbf{73.3} & 70.8/71.1 & 78.7/78.9  \\
		
		&\textbf{Ours}
		&\textbf{86.3}/\textbf{86.3}
		&\textbf{85.0}/\textbf{85.1}
		&81.9/81.9
		&\textbf{80.7}/\textbf{80.8}
		&\textbf{77.9}/\textbf{78.0}
		&76.1/76.2
		&72.2/72.2
		&\textbf{71.1}/\textbf{71.3}
		&\textbf{78.9}/\textbf{79.0}  \\
		
		\hline
		\multirow{6}{*}{\begin{tabular}[c]{@{}c@{}}CMU-MOSEI\end{tabular}}	
		
		&DCCA$^\star$  \cite{andrew2013deep}  & 80.9/80.7 & 77.3/77.4 & 74.0/73.8 & 71.2/71.1 & 69.4/69.5 & 65.4/67.5 & 63.1/66.2 & 61.0/65.6 & 70.3/71.5 \\
		&DCCAE$^\star$ \cite{wang2015deep}    & 81.2/81.2 & 78.3/78.4 & 75.4/75.5 & 72.2/72.3 & 70.0/70.3 & 66.4/69.2 & 63.2/67.6 & 62.6/66.6 & 71.2/72.6 \\
		&MCTN$^\star$  \cite{pham2019found}   & 84.2/84.2 & 81.6/81.8 & 78.7/79.0 & 76.2/76.9 & 74.1/74.3 & 72.6/73.6 & 71.1/73.2 & 70.5/72.7 & 76.1/77.0 \\
		&MMIN$^\star$  \cite{zhao2021missing} & 84.2/84.3 & 81.3/81.9 & 78.8/79.8 & 75.5/77.2 & 72.6/75.2 & 70.7/73.9 & 70.3/73.2 & 69.5/73.1 & 75.4/77.3 \\
		&GCNet$^\star$ \cite{lian2023gcnet}	  & 85.1/85.2 & 82.1/82.3 & 79.9/80.3 & 76.8/77.5 & 74.9/76.0 & 73.2/74.9 & 72.1/74.1 & 70.4/73.2 & 76.8/77.9 \\
		&DiCMoR$^\star$\cite{Wang_2023_ICCV}  & 85.1/85.1 & 83.5/83.7 & 81.5/81.8 & 79.3/79.8 & 77.4/78.7 & 75.8/77.7 & 73.7/76.7 & 72.2/75.4 & 78.6/79.9 \\
		
		&\textbf{Ours}
		&\textbf{87.3}/\textbf{87.4}
		&\textbf{86.7}/\textbf{86.8}
		&\textbf{85.7}/\textbf{85.9}
		&\textbf{84.7}/\textbf{84.8}
		&\textbf{83.8}/\textbf{84.0}
		&\textbf{82.6}/\textbf{82.8}
		&\textbf{81.3}/\textbf{81.6}
		&\textbf{80.8}/\textbf{81.0}
		&\textbf{84.1}/\textbf{84.3}   \\
		\hline
		\hline
		\text{$^\star$results come from \cite{Wang_2023_ICCV}}
	\end{tabular}
\end{table*}

\begin{table*}
	\centering
	\renewcommand\tabcolsep{10.8pt}
	\renewcommand\arraystretch{1.20}
	\caption{Comparison of performance with various missing rates on IEMOCAP. We report WAF1 scores(\%). The best performance is highlighted in bold.}
	\label{Table4}
	\begin{tabular}{cccccccccccc}
		\hline
		\multirow{2}{*}{Dataset}													&
		\multirow{2}{*}{Method}														&
		\multicolumn{8}{c}{\begin{tabular}[c]{@{}c@{}}Missing Rate\end{tabular}}	\\ \cline{3-10}
		& & 0.0	&0.1 &0.2 &0.3 &0.4 &0.5 & 0.6 & 0.7  \\
		\hline \hline

		\multirow{5}{*}{\begin{tabular}[c]{@{}c@{}}IEMOCAP \\ (four-class)\end{tabular}}

		&\textbf{SDR-GNN}   &\textbf{79.58} &78.55 &\textbf{78.08} &\textbf{77.53} &\textbf{77.09}  &\textbf{75.84} &\textbf{75.03} &\textbf{74.41} \\
		&SDR-GNN\textsubscript{w/o Sp}		 & 79.05 & 78.64& 77.72 & 76.30 & 76.95 & 75.61 & 73.70 & 73.40  \\
		&SDR-GNN\textsubscript{w/o Co}     	 & 79.26 & 78.40 & 77.93 & 76.28 & 75.23 & 74.28 & 72.94 & 72.26  \\
		&SDR-GNN\textsubscript{w/o Fre}        & 78.23 & 77.70 & 76.73 & 76.16 & 75.79 & 74.16 & 72.42 & 72.17  \\
		&SDR-GNN\textsubscript{w/o Op}         & 79.20 & \textbf{78.91} & 78.74 & 78.12 & 76.66 & 76.21 & 75.14 & 73.34  \\
		
		\hline
		\multirow{5}{*}{\begin{tabular}[c]{@{}c@{}}IEMOCAP \\ (six-class)\end{tabular}}

		&\textbf{SDR-GNN}   &\textbf{61.34} &\textbf{60.86} &\textbf{59.83} &\textbf{59.49}  &\textbf{59.16}  &\textbf{57.38} &\textbf{55.51} &\textbf{55.26} \\
		
		&SDR-GNN\textsubscript{w/o Sp}		 & 61.08 & 59.76 & 59.54 & 59.40 & 59.13 & 57.12 & 54.95 & 54.63  \\
		&SDR-GNN\textsubscript{w/o Co}     	 & 60.62 & 60.53 & 58.83 & 58.74 & 57.21 & 55.78 & 53.71 & 53.11  \\
		&SDR-GNN\textsubscript{w/o Fre}        & 59.42 & 59.21 & 59.01 & 57.17 & 56.51 & 54.82 & 53.05 & 51.97  \\
		&SDR-GNN\textsubscript{w/o Op}         & 59.58 & 60.38 & 59.01 & 58.38 & 56.55 & 55.77 & 54.38 & 53.19  \\

		\hline
		\hline
	\end{tabular}
\end{table*}

\section{Results and analysis}

\subsection{Classification Performance}
Table \ref{Table2} and Table \ref{Table3} presents the classification performance compared with different approaches under various missing rate. From these results, we can observe:

1. On average, SDR-GNN consistently outperforms other methods across all datasets. For IEMOCAP and CMU-MOSEI, SDR-GNN shows an absolute improvement from 0.77\% to 8.6\% on WAF1. Compared with non-reconstructive approaches, reconstructive techniques, including our SDR-GNN, demonstrate superior performance. This improvement is attributed to the ability of reconstructive methods to estimate and rebuild modalities from existing modalities. Compared with the reconstruction methods \cite{pham2019found,Wang_2023_ICCV,lian2023gcnet,zhao2021missing}, our SDR-GNN perform better. We argue that these baselines do not use the multi-frequency information in conversation. Our method utilizes multi-frequency signals to reconstruct missing modalities, resulting in better classification performance.

2. Our method exhibits less performance degradation with increasing missing rates compared to others. For example, in IEMOCAP (four-class), while other methods see performance drops between 6.98\% and 37.70\% as the missing rate increases to 0.7, our SDR-GNN declines by only 5.17\%. Moreover, SDR-GNN shows greater improvement as the missing rate rises; in CMU-MOSEI, the improvement is 3.21\% at a missing rate of 0.1, reaching 8.6\% at 0.7, indicating robustness in scenarios with high missing rates.

3.  Experimental results demonstrate SDR-GNN also exhibits better performance when multimodal data is complete ($\mathcal{M}=0.0$), For all datasets, our SDR-GNN improve $0.7\% \sim 2.7\%$. These results validate the effectiveness of our method on both complete and incomplete multimodal data.

\subsection{Ablation Study}	
To study the necessity of different components in SDR-GNN to model performances, we conduct ablation studies on IEMOCAP(four-class) and IEMOCAP(six-class). Experimental results are shown in Table \ref{Table4}.

\begin{itemize}
	
	\item SDR-GNN: Our proposed method that considers both features relationships and multi-frequency information.
	
	\item SDR-GNN\textsubscript{w/o Sp}: It is derived from SDR-GNN, but ignores the information comes from spearker interaction graph.
	
	\item SDR-GNN\textsubscript{w/o Co}: It is derived from SDR-GNN, but ignores the information comes from context interaction graph.
	
	\item SDR-GNN\textsubscript{w/o Fre}: It is derived from SDR-GNN, but replaces the frequency-aware graph learning with a GNN-based model from DialogueGCN \cite{ghosal2019dialoguegcn}, a currently advanced graphical model for conversation understanding.
	
	\item SDR-GNN\textsubscript{w/o Op}: It is derived from SDR-GNN, but remove self optimizing multi-head attention layer used for data reconstruction.
	
\end{itemize}

\textbf{Impact of Speaker Interaction Graph:}
To study the effect of speaker interaction graph. We remove the information that comes from the speaker graph. Experimental results show that performance decreases in most cases on both IEMOCAP (four-class) and IEMOCAP (six-class). The inferior performance of SDR-GNN\textsubscript{w/o Sp} on both datasets proves the effectiveness of speaker information.

\textbf{Impact of Context Interaction Graph:}
We remove the information that comes from the context interaction graph to investigate its effectiveness. Experimental results show that performance decreases at all missing rates. Meanwhile, compared with SDR-GNN\textsubscript{w/o Sp}, SDR-GNN\textsubscript{w/o Co} deceases more. The results show that contextual information is more important than speaker information, which also proves the significance of contextual information.

\textbf{Impact of Frequency-aware Graph Learning:}
To investigate the impact of frequency-aware graph learning, we replace the frequency-aware graph learning with a graph convolution network from DialogueGCN, which captures speaker and context dependencies on one graph. Results from Table \ref{Table4} show that performance drop considerably at all missing rates. This proves the importance and superiority of capturing frequency information using Frequency-aware Graph Learning, especially when modalities are incomplete.

\textbf{Impact of Self Optimization:}
We use multi-head attention layers to optimize the reconstructed data. To study the effect of this model, we remove the multi-head attention layers during training. Experimental results demonstrate that the performances of SDR-GNN\textsubscript{w/o Op} decline on all datasets. The results of SDR-GNN\textsubscript{w/o Op} prove the effectiveness of self optimizing reconstructed data.

\begin{figure*}[!b]
	\centering
	\subfloat[$\mathcal{M}$ = 0.1]{
		\includegraphics[width=0.22\linewidth, trim=42 42 42 42]{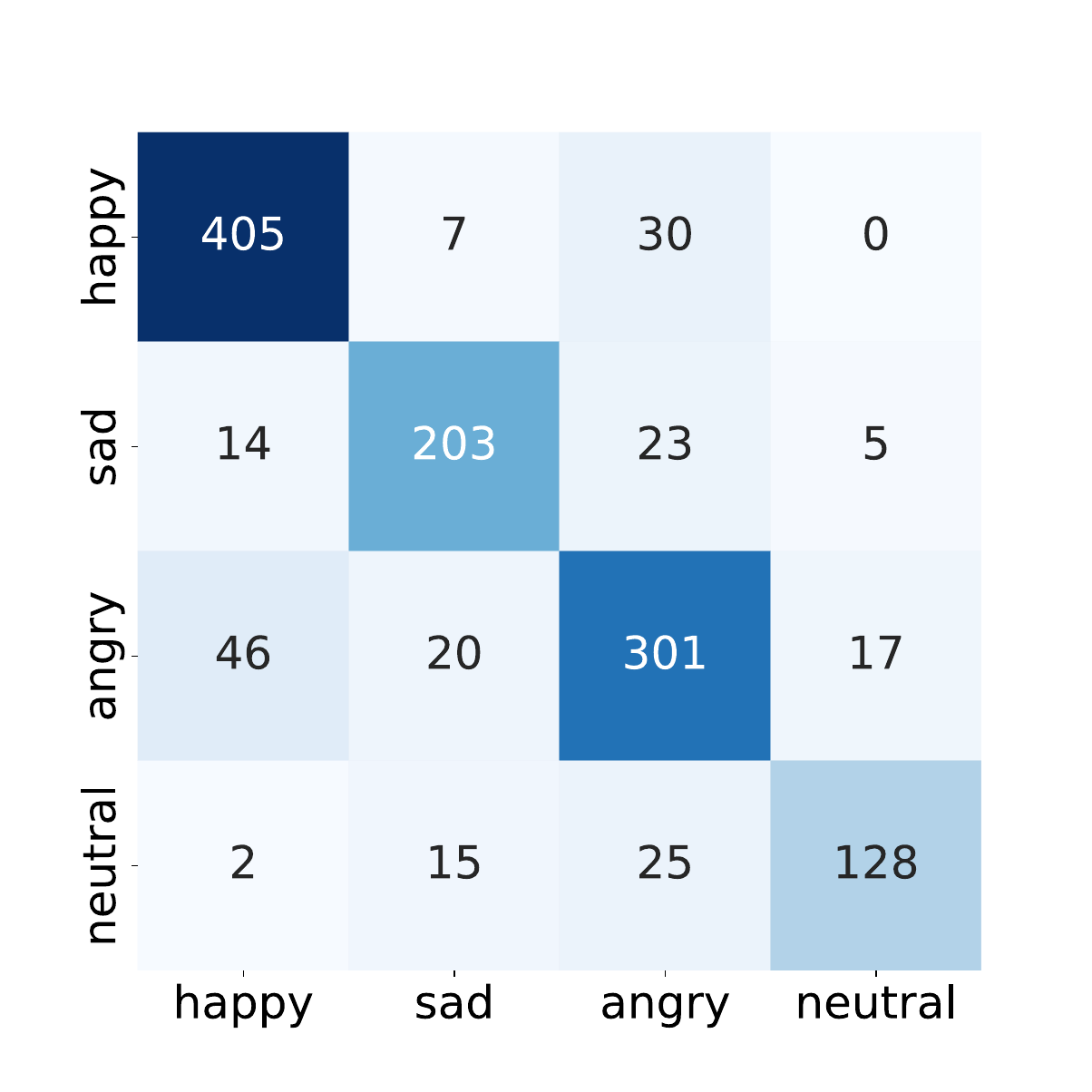}
		\label{Figure3-1}
	}
	\hfill
	\subfloat[$\mathcal{M}$ = 0.3]{
		\includegraphics[width=0.22\linewidth, trim=42 42 42 42]{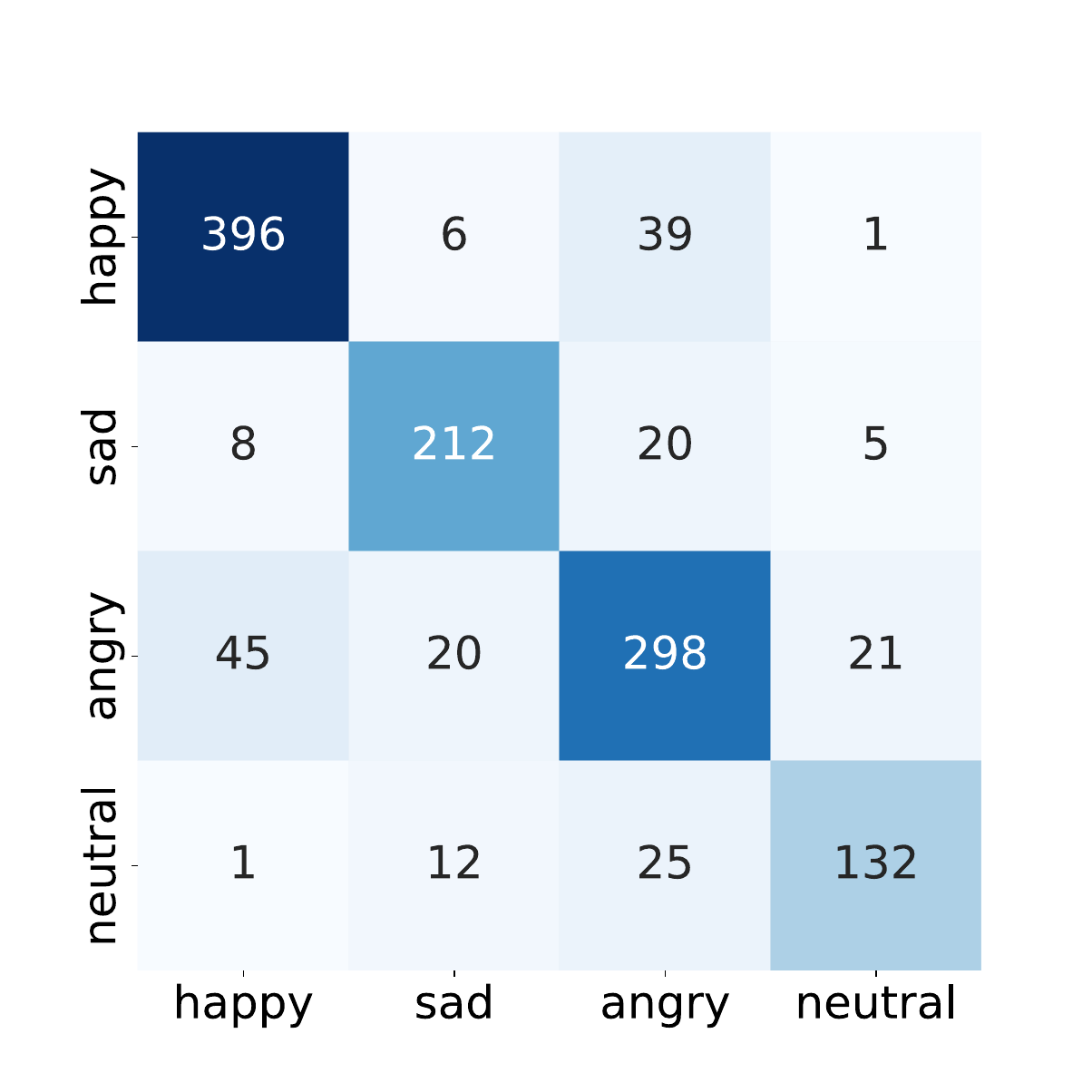}
		\label{Figure3-2}
	}
	\hfill
	\subfloat[$\mathcal{M}$ = 0.5]{
		\includegraphics[width=0.22\linewidth, trim=42 42 42 42]{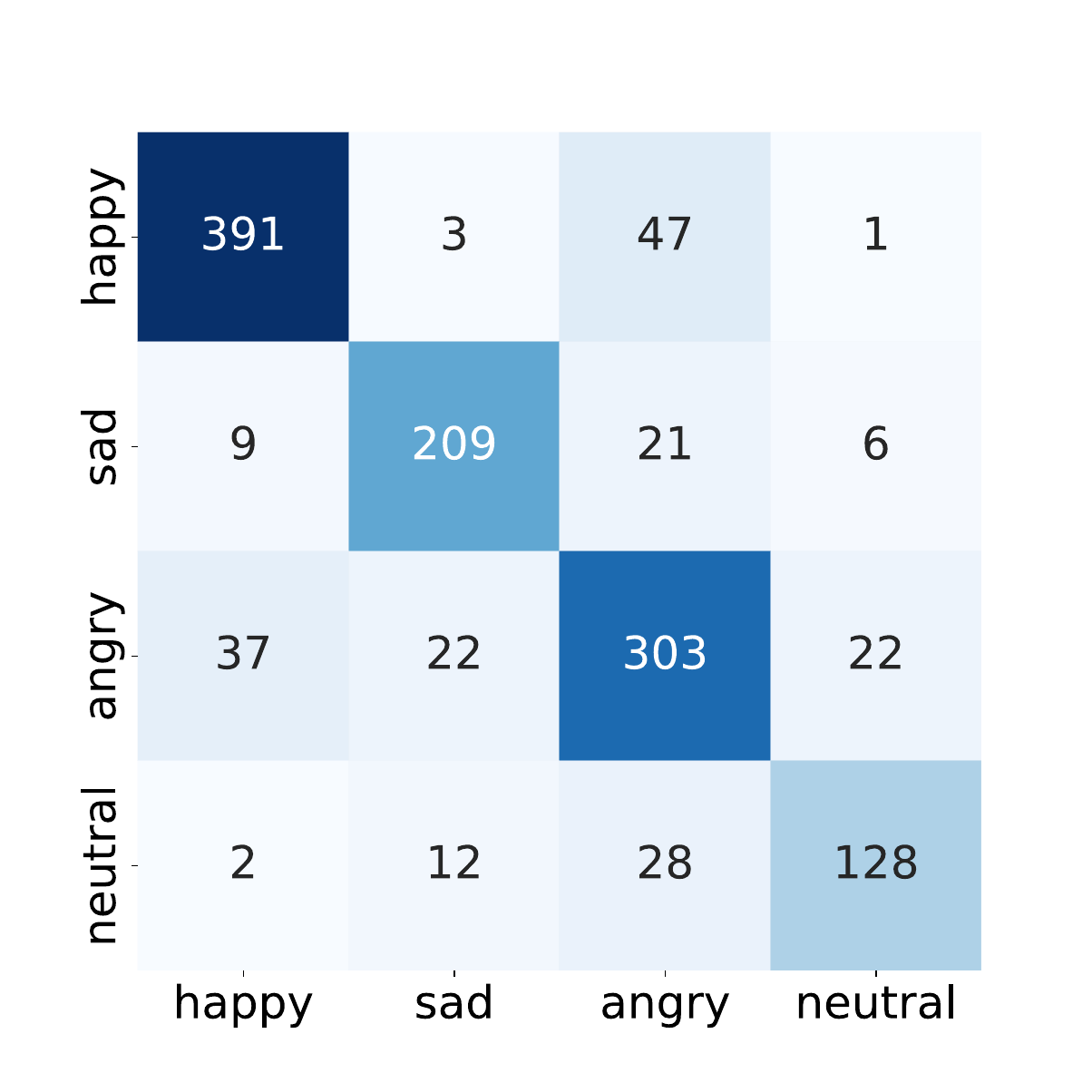}
		\label{Figure3-3}
	}
	\hfill
	\subfloat[$\mathcal{M}$ = 0.7]{
		\includegraphics[width=0.22\linewidth, trim=42 42 42 42]{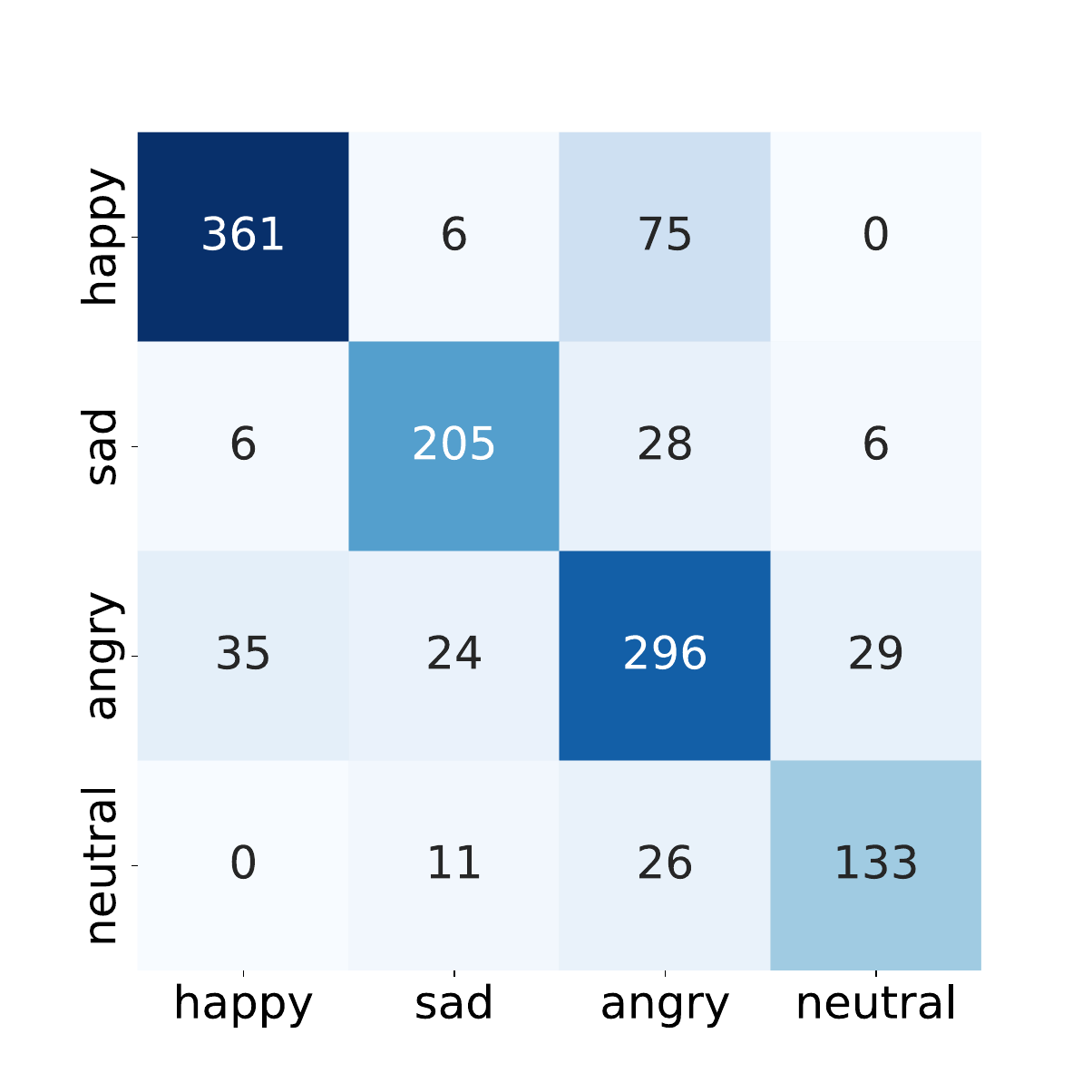}
		\label{Figure3-4}
	}
	
	\vspace{0.7cm} 
	\subfloat[$\mathcal{M}$ = 0.1]{
		\includegraphics[width=0.22\linewidth, trim=42 42 42 42]{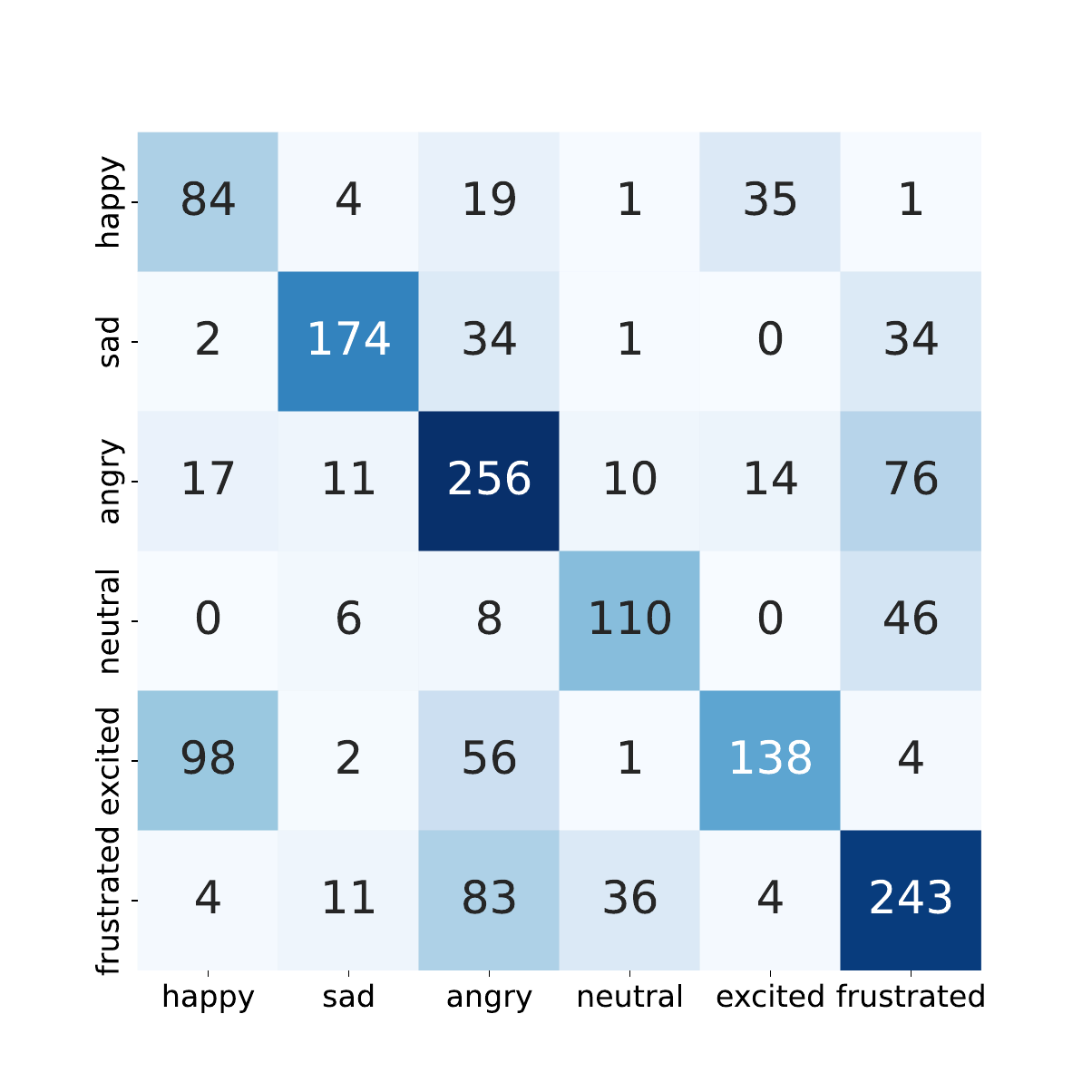}
		\label{Figure3-5}
	}
	\hfill
	\subfloat[$\mathcal{M}$ = 0.3]{
		\includegraphics[width=0.22\linewidth, trim=42 42 42 42]{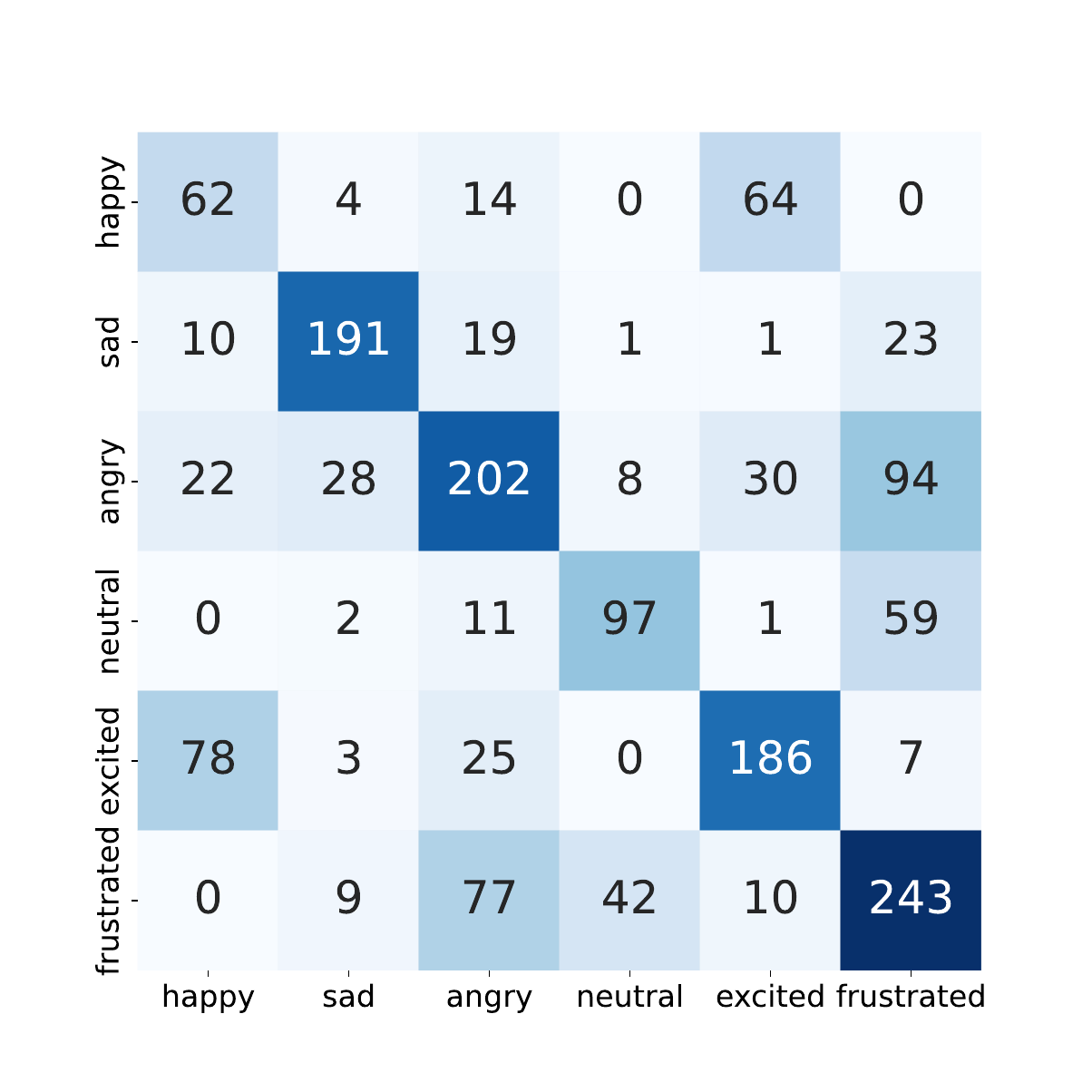}
		\label{Figure3-6}
	}
	\hfill
	\subfloat[$\mathcal{M}$ = 0.5]{
		\includegraphics[width=0.22\linewidth, trim=42 42 42 42]{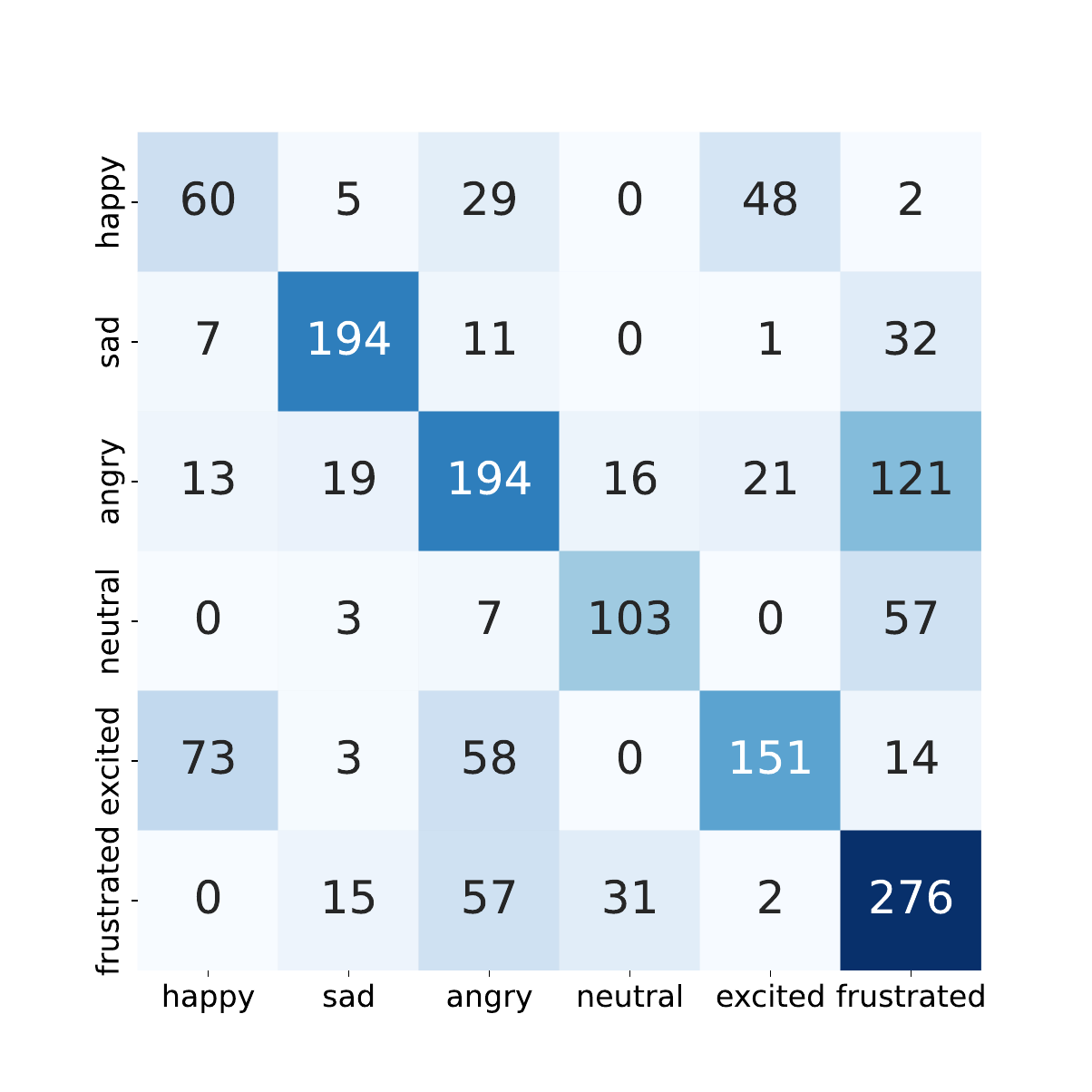}
		\label{Figure3-7}
	}
	\hfill
	\subfloat[$\mathcal{M}$ = 0.7]{
		\includegraphics[width=0.23\linewidth, trim=42 42 42 42]{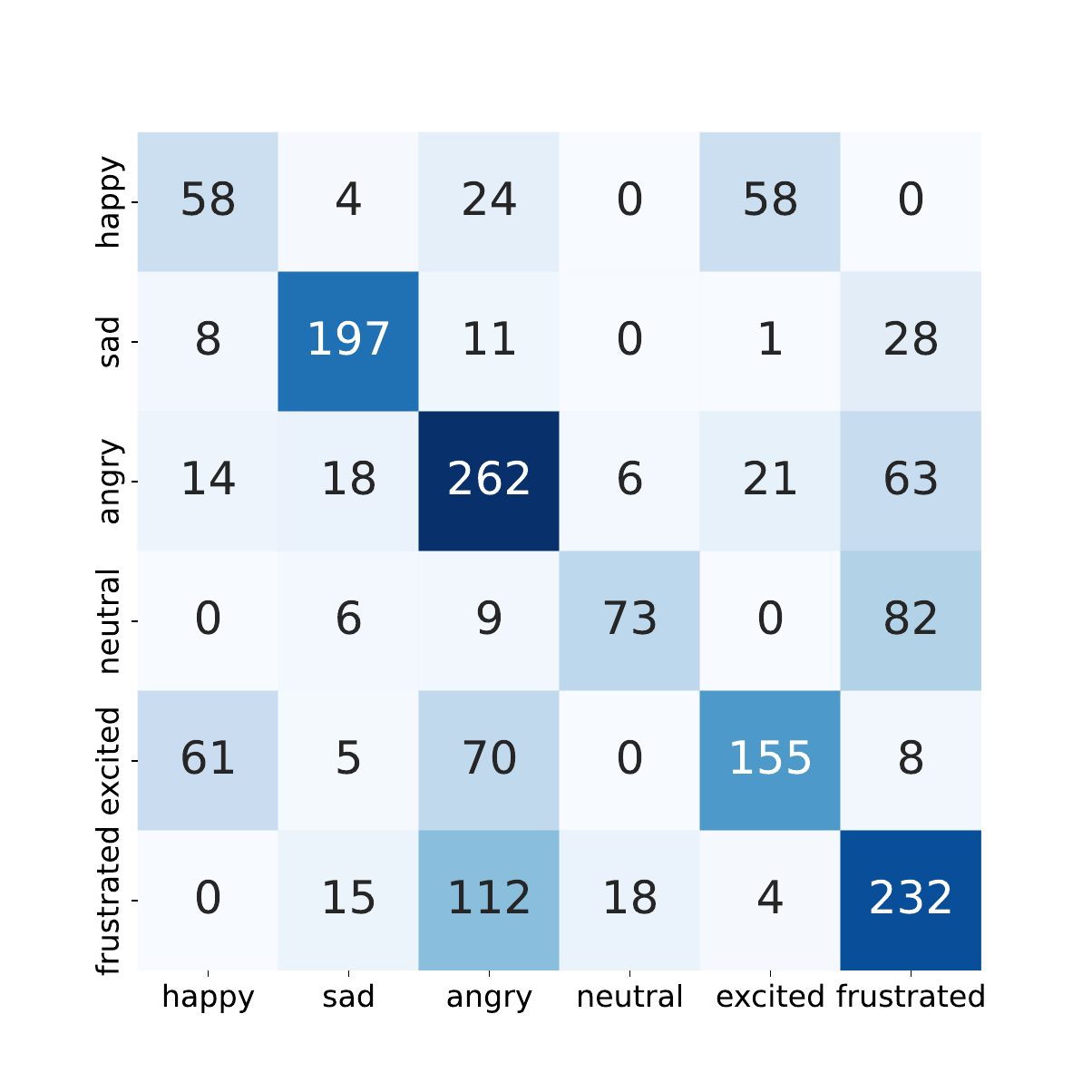}
		\label{Figure3-8}
	}
	\caption{Confusion matrices of the test set on IEMOCAP at varying missing rates. The matrices present the true labels along its rows and the predicted labels across its columns.}
	\label{Figure3}
\end{figure*}

\begin{figure*}[t]
	\centering
	\subfloat[IEMOCAP(four-class)]{
		\includegraphics[width=0.45\linewidth]{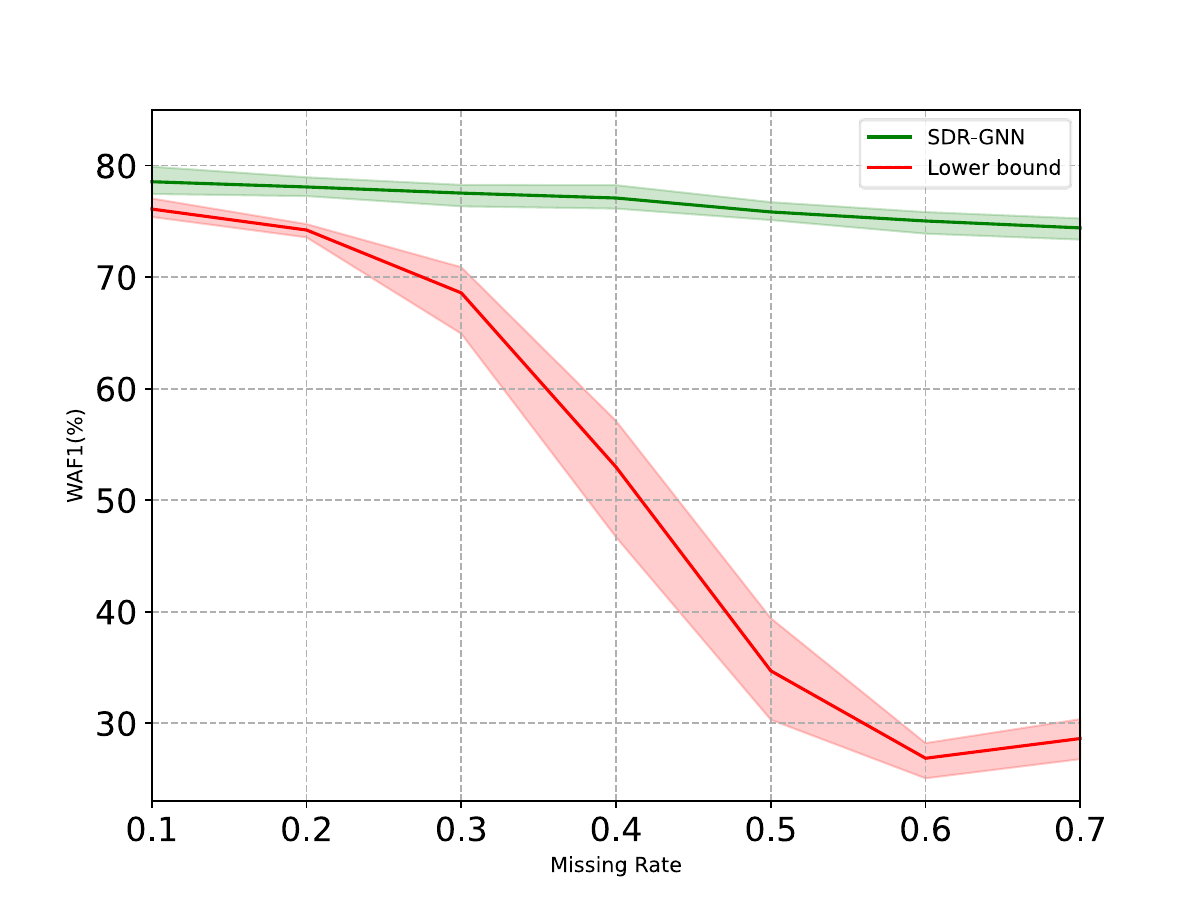}
		\label{Figure4-1}
	}
	\subfloat[IEMOCAP(six-class)]{
		\includegraphics[width=0.45\linewidth]{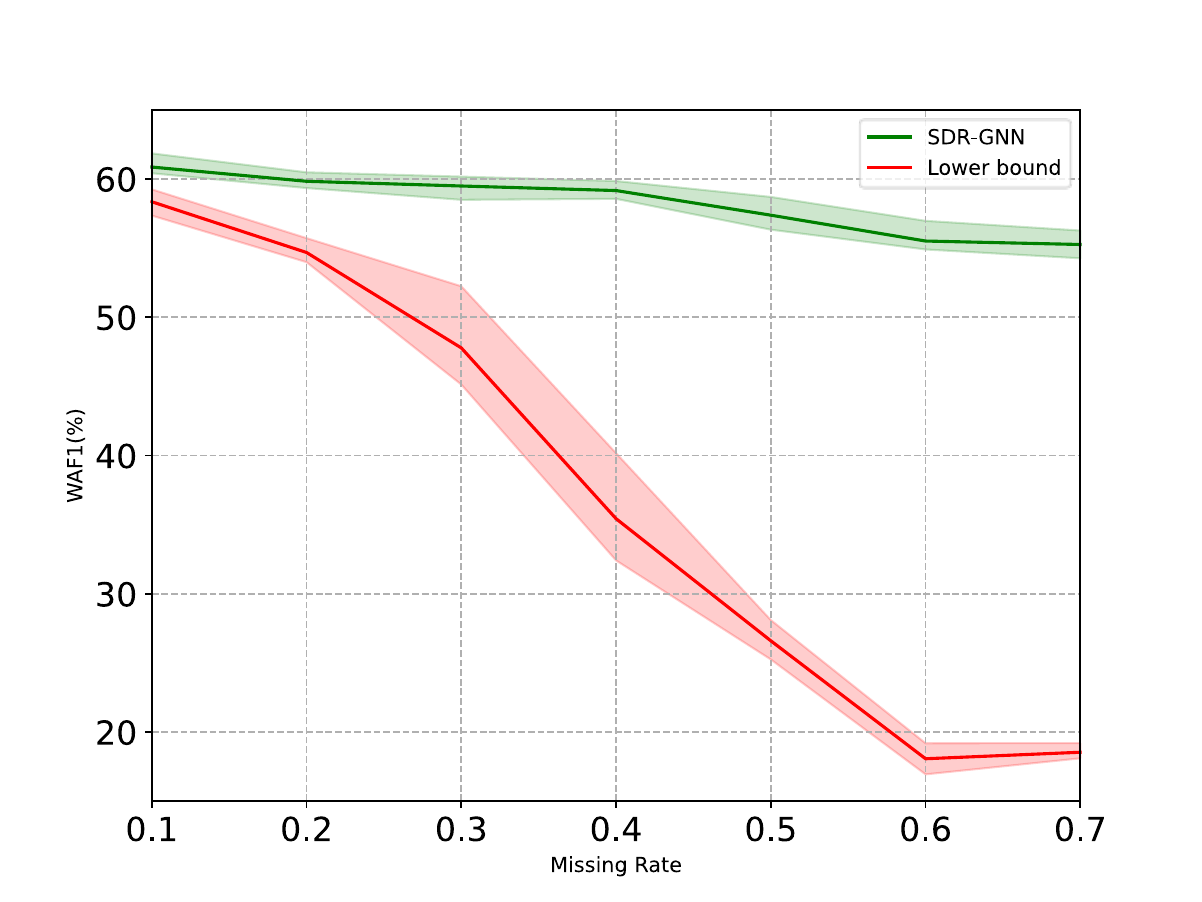}
		\label{Figure4-2}
	}
	
	\subfloat[CMU-MOSI]{
		\includegraphics[width=0.45\linewidth]{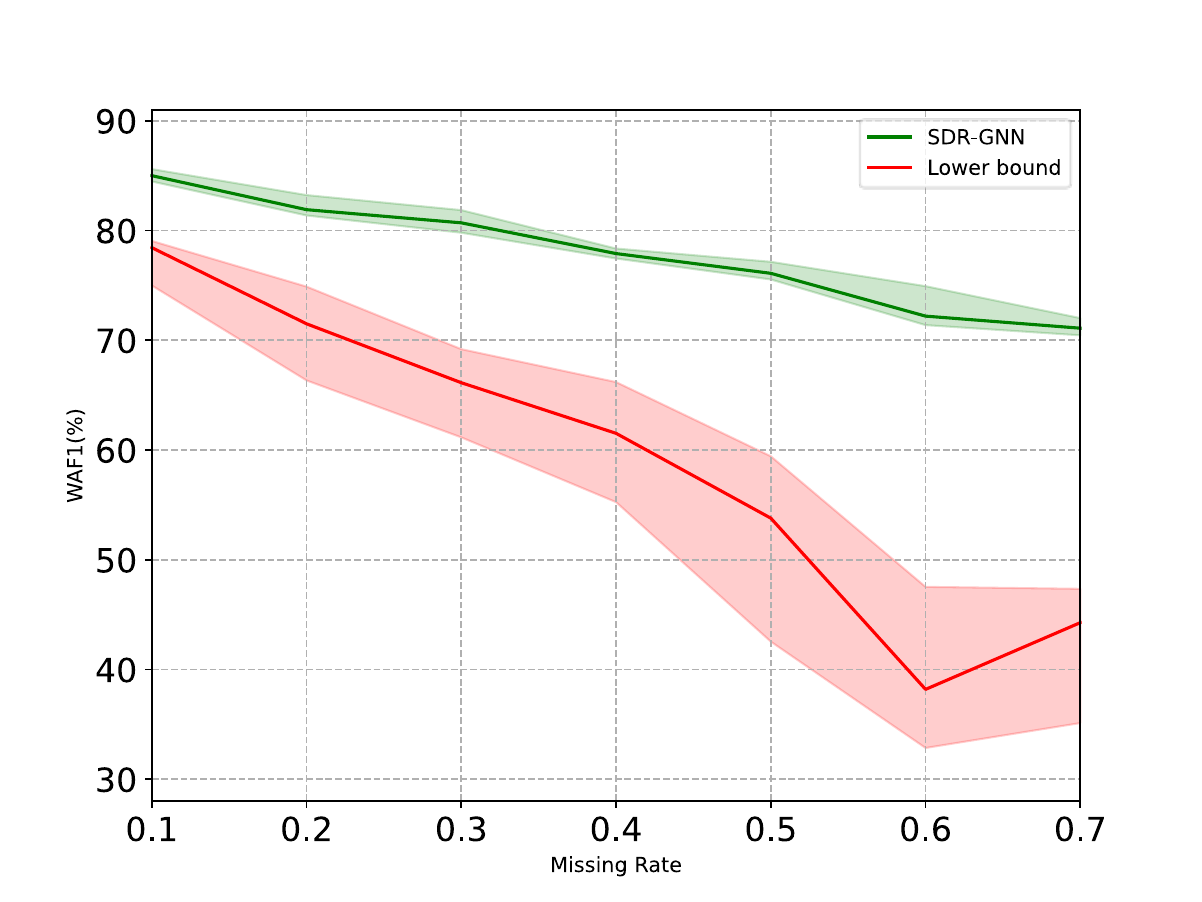}
		\label{Figure4-3}
	}
	\subfloat[CMU-MOSEI]{
		\includegraphics[width=0.45\linewidth]{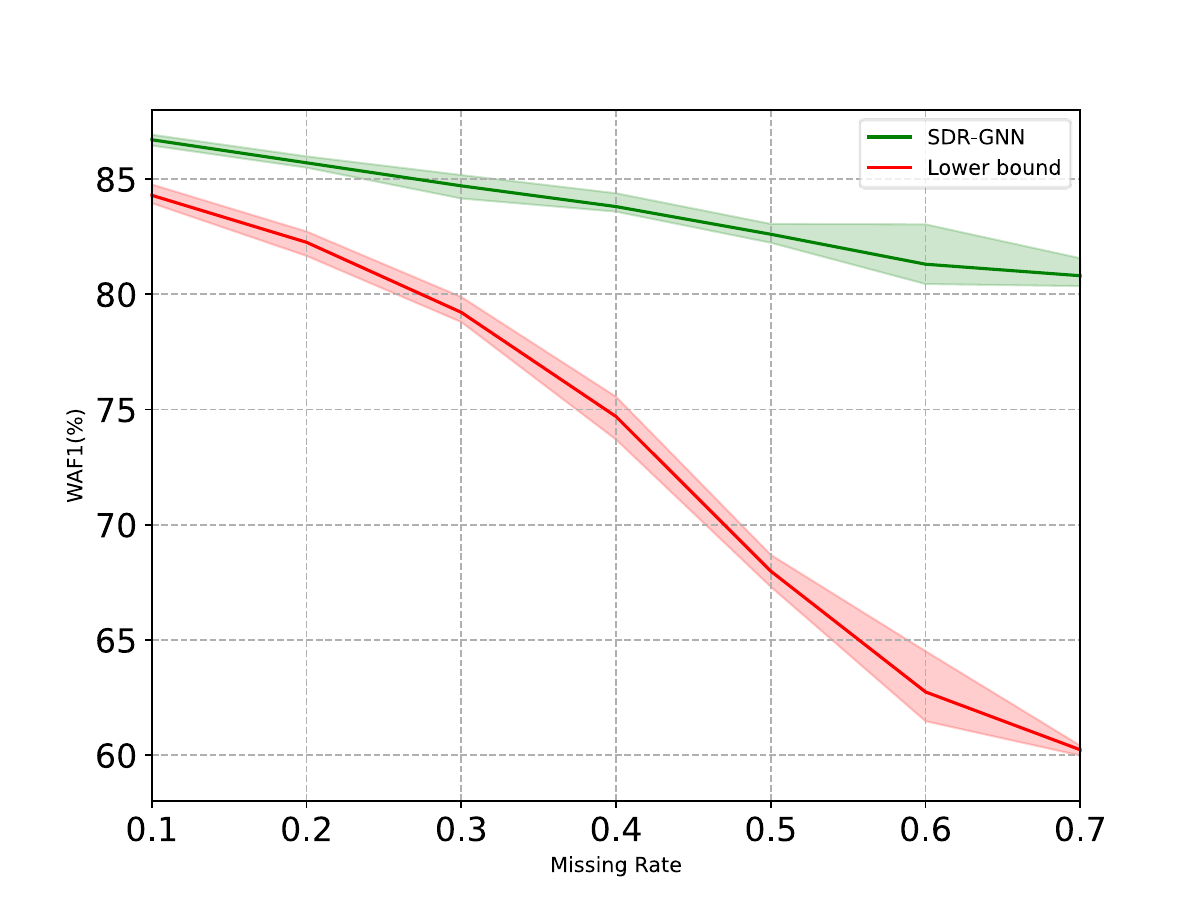}
		\label{Figure4-4}
	}
	
	\caption{Classification performance comparison between SDR-GNN and Lower bound under different missing rates.}
	\label{Figure4}
\end{figure*}

\subsection{Emotion Categories Analysis}
We investigate the classification performance of different emotional categories under various missing rates. Fig. \ref{Figure3-1}$\sim$\ref{Figure3-8} show confusion matrices on IEMOCAP (four-class) and IEMOCAP (six-class) under different missing rates. The rows represent the predicted labels, and the columns represent the actual emotional labels.

Fig. \ref{Figure3-1}$\sim$\ref{Figure3-4} depict the confusion matrices on IEMOCAP (four-class). From these matrices, we observe no significant decrease in the accuracy of recognizing various emotion categories as the missing rate increases. This indicates that our SDR-GNN can effectively recognize conversations with high missing rates. However, as the missing rate increases, we notice that conversations truly labeled as `happy' are more likely to be misclassified as `angry'. We attribute this to the possibility that, with significant data loss, the model may struggle to capture subtle features distinguishing between happy and angry emotions. For instance, incomplete tone and emphasis information in audio may hinder the model's ability to differentiate between excited high tones and angry high tones. In the expression of happy and angry emotions, certain expressions may appear similar to some extent, particularly when multimodal information is incomplete. Without contextual support from other modalities, the model may fail to interpret these subtle differences accurately.

Fig. \ref{Figure3-5}$\sim$\ref{Figure3-8} depict the confusion matrix on IEMOCAP (six-class). Similar to IEMOCAP (four-class), the model can maintain recognition accuracy even as the loss rate increases. However, unlike IEMOCAP (four-class), IEMOCAP (six-class) introduces two additional labels: "excited" and "frustrated," which adds complexity to the model's recognition task.
From these confusion matrices, it's evident that statements labeled as "happy" are more prone to being misclassified as "angry" or "excited." This is reasonable since the model struggles to differentiate statements with intense emotions, particularly when the loss rate is high.
Moreover, statements with a true label of "neutral" are often mistakenly identified as "frustrated". From a frequency perspective, we believe that emotions such as "neutral" exhibit low-frequency signals that tend towards zero. However, when the loss rate increases, some low-frequency signals are erroneously amplified, leading to misjudgments by the model.
Additionally, "frustrated" is frequently misclassified as "anger," likely due to similarities in language features between frustrated and anger, such as negative emotions and vocabulary. This similarity poses challenges in accurately distinguishing between these two emotions.

\textcolor{black}{\subsection{Model Complexity Analysis}}

\textcolor{black}{
	To analyze the complexity of our model, we compared SDR-GNN and SDR-GNNmini with other state-of-the-art models.}

\begin{itemize}
	\color{black}
	
	\item SDR-GNN: Our original version that considers both features relationships and multi-frequency information..
	
	\item SDR-GNN\textsubscript{mini}: A derivative of SDR-GNN that retains all core functionalities of SDR-GNN but reduces the number of neurons and network layers.
	
\end{itemize}

\textcolor{black}{From the experimental results in Table \ref{Table11}, SDR-GNN performs the best, but its parameter size and training speed are inferior to other methods. SDR-GNN\textsubscript{mini} outperforms other methods in terms of parameter size and training speed, but its performance is slightly lower than SDR-GNN.}

\textcolor{black}{We believe that the higher parameter size of SDR-GNN enhances the model's learning capacity, thereby improving its performance, but this also results in longer training times. SDR-GNN\textsubscript{mini}, on the other hand, sacrifices some performance in exchange for faster training speed.}

\textcolor{black}{In conclusion, SDR-GNN\textsubscript{mini} outperforms other solutions in terms of parameter size, training time, and performance, which also validates the effectiveness of our method.}

\begin{table}[h]
	\centering
	\color{black}
	\renewcommand\tabcolsep{5.0pt}
	\renewcommand\arraystretch{1.20}
	\caption{\textcolor{black}{The complexity comparison of different models on IEMOCAP(four-class) under missing rate $\mathcal{M}$ = 0.6. We report Parameters(M), Training Time(s) and WAF1 scores(\%). The best performance is highlighted in bold.}}
	\label{Table11}
	\begin{tabular}{cc|cccc}
		\hline
		\multicolumn{2}{c|}{Models} & {Params(M)} &{Training Time(s)} &{WAF1(\%)}\\
		\hline
		\hline

		\multicolumn{2}{c|}{CPM-Net} & 37.7  & 8.34 & 68.68	 \\

		\multicolumn{2}{c|}{GCNet}  & 34.0  & 7.68 & 78.87   \\
		\hline
		
		\multicolumn{2}{c|}{SDR-GNN}     & 41.1  & 10.67 & \textbf{81.13}  	\\

		\multicolumn{2}{c|}{SDR-GNN\textsubscript{mini}} 	 & \textbf{32.7}  & \textbf{7.52}   & 80.34  \\
		\hline
		\hline
	\end{tabular}
\end{table}

\subsection{Importance of Incomplete Data}
Our proposed SDR-GNN not only utilizes complete multimodal data, but also make full use of incomplete multimodal data. To investigate the importance of incomplete data, in Fig. \ref{Figure3}, we compare the performance of different methods under various missing rates.

\begin{itemize}
	
	\item SDR-GNN: The method we proposed that fully utilizes both complete and incomplete modality data for conversational learning.
	
	\item Lower bound: It comes from SDR-GNN, but abandons the incomplete multimodal utterances. This method is a straightforward strategy that only focus on complete data, which is regarded as the lower bound \cite{ma2021maximum}.
	
\end{itemize}

According to Fig. \ref{Figure3}, SDR-GNN consistently outperforms the lower bound across all missing rates and datasets. Meanwhile, as the missing rate increases, the disparity in performance between SDR-GNN and the comparison system widens significantly.This observation underscores the significance of leveraging incomplete data to enhance the performance of conversational learning models. By effectively incorporating incomplete information, SDR-GNN demonstrates superior adaptability and robustness in handling incomplete multimodal data.

The experimental results demonstrate that despite the incompleteness of the data modality, it retains significant utility. It is imperative to concurrently leverage both complete and incomplete modal data to enhance contextual understanding and improve recognition outcomes. Our belief stems from the comprehensive utilization of both data types by SDR-GNN in establishing contextual connections, enabling it to maintain recognition accuracy even under high missing rates. This underscores the efficacy and superiority of leveraging both data types simultaneously.

\begin{figure*}[t]
	\centering
	\subfloat[IEMOCAP(four-class)]{
		\includegraphics[width=0.45\linewidth]{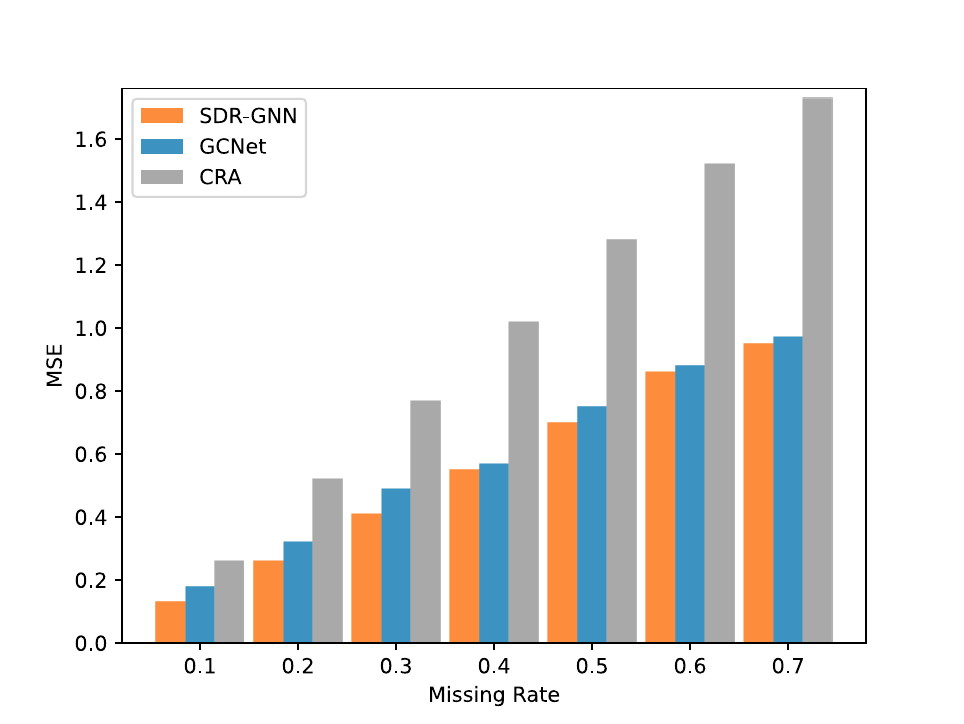}
		\label{Figure5-1}
	}
	\subfloat[IEMOCAP(six-class)]{
		\includegraphics[width=0.45\linewidth]{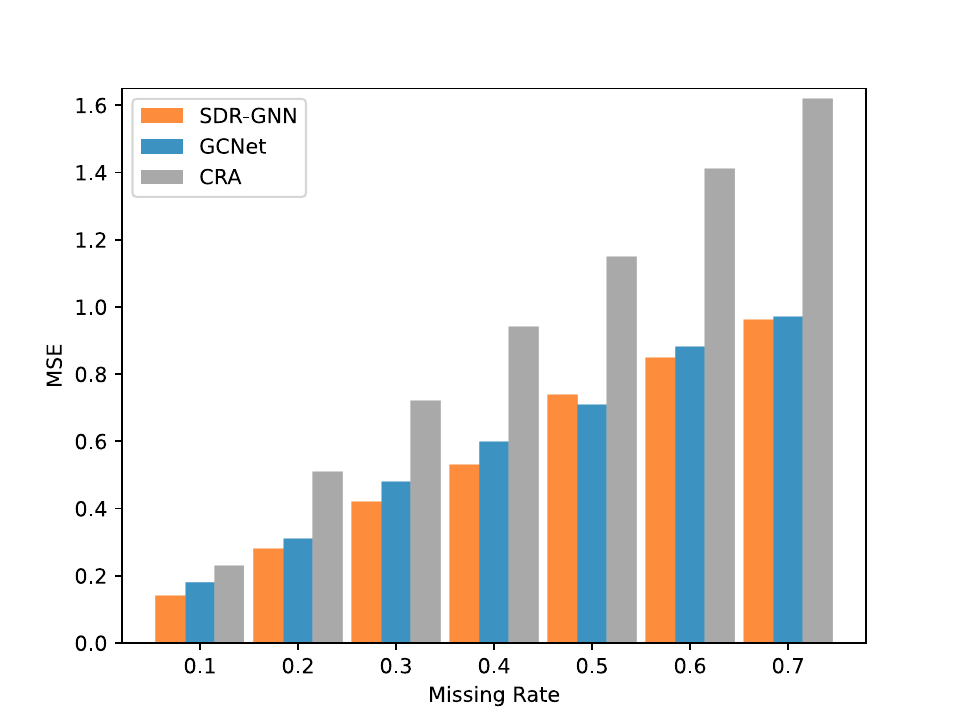}
		\label{Figure5-2}
	}
	
	\subfloat[CMU-MOSI]{
		\includegraphics[width=0.45\linewidth]{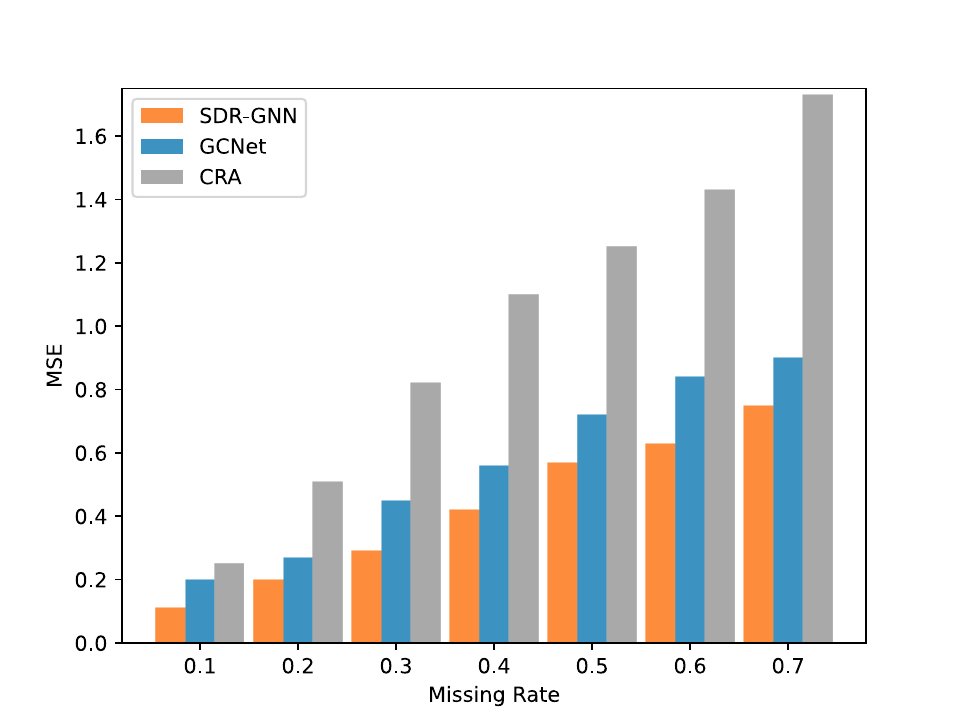}
		\label{Figure5-3}
	}
	\subfloat[CMU-MOSEI]{
		\includegraphics[width=0.45\linewidth]{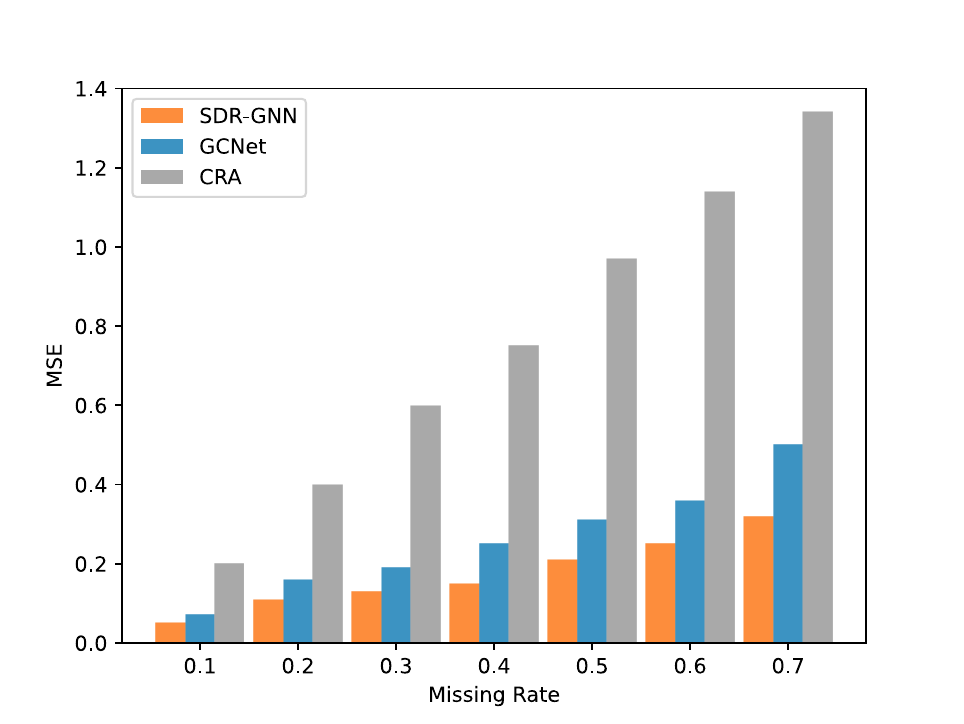}
		\label{Figure5-4}
	}
	
	\caption{\textcolor{black}{Reconstruction performance comparison between SDR-GNN and other methods under different missing rates. Lower MSE indicates better imputation performance.}}
	\label{Figure5}
\end{figure*}

\textcolor{black}{\subsection{Reconstruction Performance}}

\textcolor{black}{Our approach employs SDR-GNN to reconstruct the data in order to meet the requirements of downstream emotion classification. Therefore, the quality of the reconstructed data will directly impact the performance of the classification task. To validate the effectiveness of our approach, we compared it with two advanced data reconstruction models, GCNet \cite{lian2023gcnet} and CRA \cite{tran2017missing}. To evaluate the reconstruction performance of different methods, we calculated the mean square error (MSE) between the reconstructed data of the missing modalities and the real data, in line with previous works.}

\textcolor{black}{Fig. \ref{Figure5} shows the performance of the reconstructed data under different missing rates. A lower MSE indicates a smaller difference between the reconstructed data and the real data, implying better reconstruction performance. 
	We observe that as the missing rate increases, the MSE also increases. This is because a higher missing rate leads to a reduction in data volume, making it more difficult for the model to reconstruct the data. }

\textcolor{black}{The experimental results demonstrate that SDR-GNN outperforms other methods in most cases. Compared to GNN-based models, SDR-GNN performs better because we utilize multi-frequency signals of different frequencies in our reconstruction method, further proving the importance of multi-frequency information for data reconstruction.
Moreover, as the missing rate increases, the growth in MSE for SDR-GNN is the lowest, indicating that our model has better robustness compared to other methods.}

\begin{table}[t]
	\centering
	\renewcommand\tabcolsep{10pt}
	\renewcommand\arraystretch{1}
	\caption{Parameter tuning with various missing rates.}
	\label{Table5}
	\begin{tabular}{c|c|cccc}
		\hline
		\multirow{2}{*}{$\mathcal{M}$}												&
		\multirow{2}{*}{$h$}														&
		\multicolumn{4}{c}{\begin{tabular}[c]{@{}c@{}}$w$\end{tabular}}
		\\ \cline{3-6}
		&& 1	& 2 & 3 & 4 \\
		\hline
		\hline
		\multirow{4}{*}{\begin{tabular}[c]{@{}c@{}}0.0\end{tabular}}		
		&100 	  	& 77.85	    & 78.02  	& 77.87  	& 78.15 	            \\
		&\textcolor{black}{150} &\textcolor{black}{78.75} &\textcolor{black}{78.55} &\textcolor{black}{79.03} &\textbf{\textcolor{black}{79.52}}\\
		&200		& 78.76  	& 79.10  	& 78.99  	& 79.45		\\
		&\textcolor{black}{250}    &\textcolor{black}{78.68} &\textcolor{black}{79.32} &\textcolor{black}{79.12} &\textcolor{black}{78.50} \\
		\hline
		\multirow{4}{*}{\begin{tabular}[c]{@{}c@{}}0.1\end{tabular}}	
		&100		& 77.95  	& 77.50  	& 77.94  	& 77.61	   		\\
		&\textcolor{black}{150}    &\textcolor{black}{77.69} &\textcolor{black}{77.75} &\textbf{\textcolor{black}{79.05}} &\textcolor{black}{78.38}\\
		&200		& 78.50	 	& 78.37  	& 78.58	& 78.57		\\
		&\textcolor{black}{250}    &\textcolor{black}{78.43} &\textcolor{black}{78.80} &\textcolor{black}{78.33} &\textcolor{black}{78.24} \\
		\hline
		\multirow{4}{*}{\begin{tabular}[c]{@{}c@{}}0.2\end{tabular}}	
		&100		& 77.26 	& 77.34 	& 77.74		& 77.21	\\
		&\textcolor{black}{150}    &\textcolor{black}{77.62} &\textcolor{black}{77.71} &\textcolor{black}{77.85} &\textcolor{black}{77.44}\\
		&200		& 77.36		& 76.67 	& \textbf{78.12}	& 78.00	\\
		&\textcolor{black}{250}    &\textcolor{black}{77.54} &\textcolor{black}{77.43} &\textcolor{black}{77.88} &\textcolor{black}{77.32}\\
		\hline
		\multirow{4}{*}{\begin{tabular}[c]{@{}c@{}}0.3\end{tabular}}
		&100		& 76.80 	& 76.63		& 77.04		& 77.04 			\\
		&\textcolor{black}{150}    &\textcolor{black}{76.96} &\textcolor{black}{77.18} &\textcolor{black}{77.32} &\textcolor{black}{76.95}\\
		&200		& 76.92		& 76.67		& \textbf{77.63} 	& 77.44		\\
		&\textcolor{black}{250}    &\textcolor{black}{76.60} &\textcolor{black}{76.03} &\textcolor{black}{76.78} &\textcolor{black}{76.66}\\
		\hline
		\multirow{4}{*}{\begin{tabular}[c]{@{}c@{}}0.4\end{tabular}}
		&100		& 75.69 	& 75.81		& 75.97		& 76.40 			\\
		&\textcolor{black}{150}    &\textcolor{black}{75.66} &\textcolor{black}{75.88} &\textcolor{black}{76.23} &\textcolor{black}{76.55}\\
		&200		& 75.93		& 76.74 	& \textbf{77.11} 	& 76.70					\\
		&\textcolor{black}{250}    &\textcolor{black}{75.98} &\textcolor{black}{75.50} &\textcolor{black}{75.65} &\textcolor{black}{76.23}\\
		\hline
		\multirow{4}{*}{\begin{tabular}[c]{@{}c@{}}0.5\end{tabular}}
		&100		& 74.99 	& 75.23 	& 75.51     & 75.64		\\
		&\textcolor{black}{150}    &\textcolor{black}{75.22} &\textcolor{black}{75.44} &\textcolor{black}{75.61} &\textcolor{black}{75.67}\\
		&200		& 75.73		& 75.32 	& 75.92     &\textbf{76.09}		\\
		&\textcolor{black}{250}    &\textcolor{black}{75.02} &\textcolor{black}{74.87} &\textcolor{black}{75.56} &\textcolor{black}{75.62}\\
		\hline
		\multirow{4}{*}{\begin{tabular}[c]{@{}c@{}}0.6\end{tabular}}
		&100		& 74.41		& 74.26     & 74.33     & 74.64  			\\
		&\textcolor{black}{150}    &\textcolor{black}{74.54} &\textcolor{black}{74.36} &\textcolor{black}{74.76} &\textcolor{black}{74.82}\\
		&200		& 74.84		& 74.74 	& \textbf{75.55}	 & 74.88 \\
		&\textcolor{black}{250}    &\textcolor{black}{74.34} &\textcolor{black}{74.02} &\textcolor{black}{73.89} &\textcolor{black}{74.44}\\
		\hline
		\multirow{4}{*}{\begin{tabular}[c]{@{}c@{}}0.7\end{tabular}}
		&100		& 74.33		& 73.90     & 74.30     & 74.18			\\
		&\textcolor{black}{150}    &\textcolor{black}{74.43} &\textcolor{black}{73.98} &\textcolor{black}{74.77} &\textcolor{black}{73.69}\\
		&200		& 74.28		& 74.09 	& \textbf{74.78}	 & 74.10 \\
		&\textcolor{black}{250}    &\textcolor{black}{72.13} &\textcolor{black}{73.34} &\textcolor{black}{73.23} &\textcolor{black}{74.02} \\
		\hline 	
		\hline
		
	\end{tabular}
\end{table}

\begin{figure}[t]
	\includegraphics[width=0.99\linewidth]{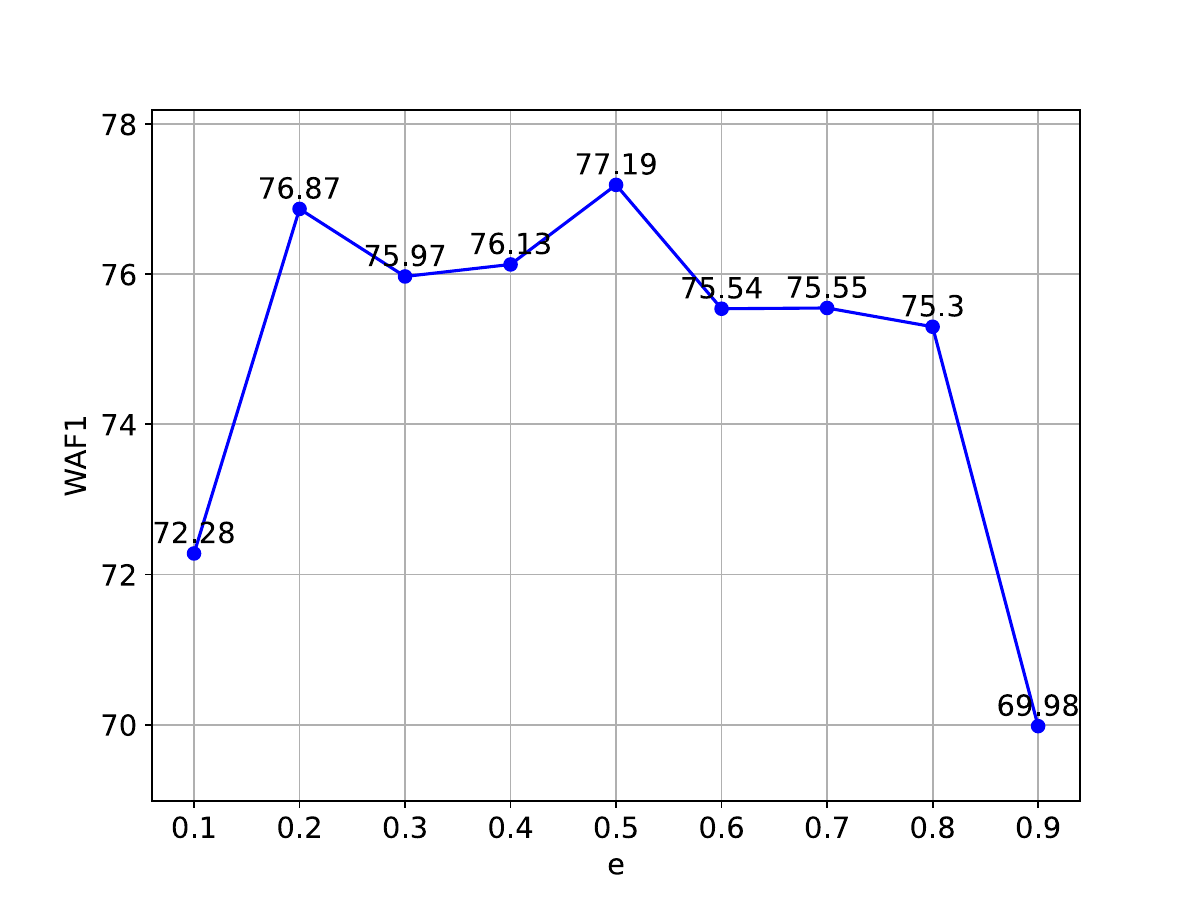}
	\caption{\textcolor{black}{Parameter tuning with various number of hypergraph from IEMOCAP(Four).}}
	\label{Figure6}
\end{figure}

\begin{figure}[t]
	\includegraphics[width=0.99\linewidth]{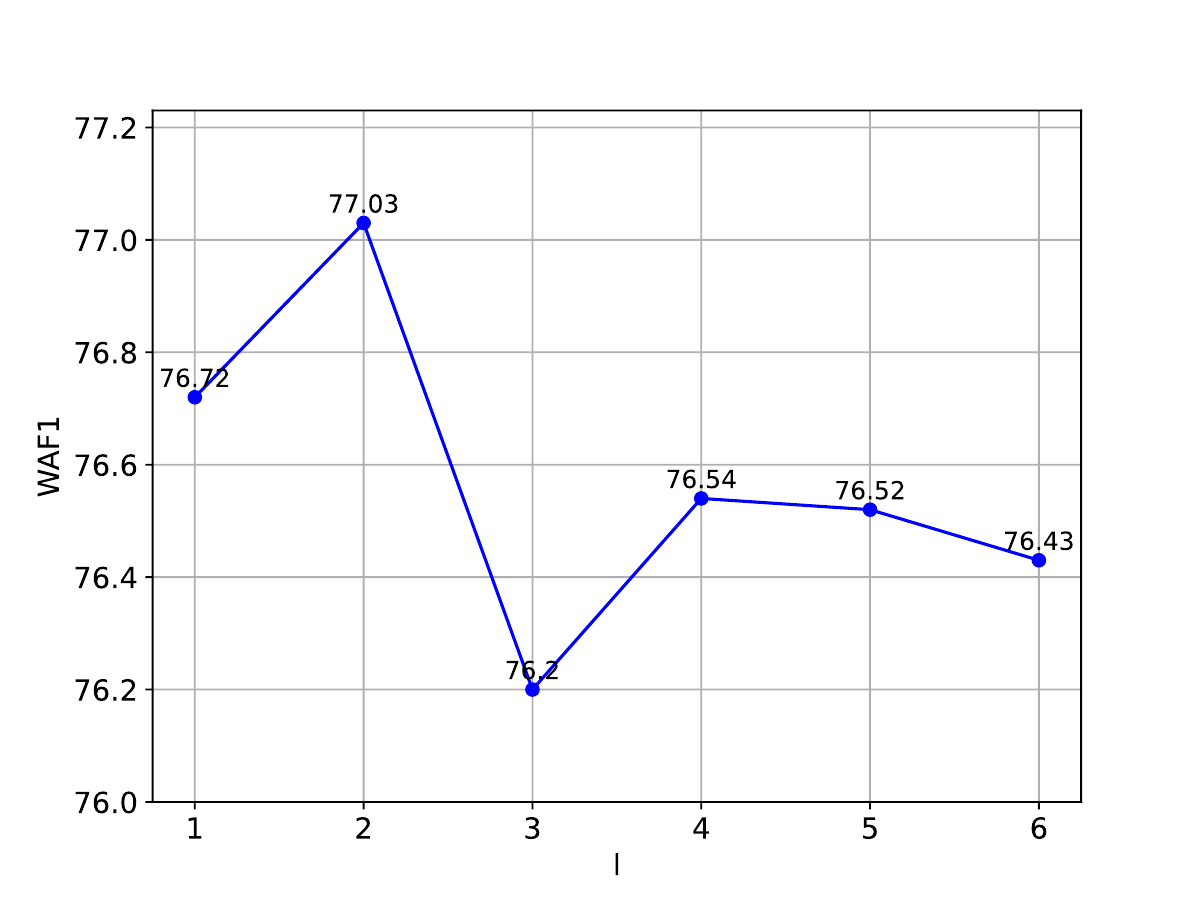}
	\caption{\textcolor{black}{Parameter tuning with various weight of reconstruction loss function from IEMOCAP (Four-class).}}
	\label{Figure7}
\end{figure}

\subsection{Parameter Tuning}
\begin{figure*}[t]
	\centering
	\includegraphics[width=0.99\linewidth]{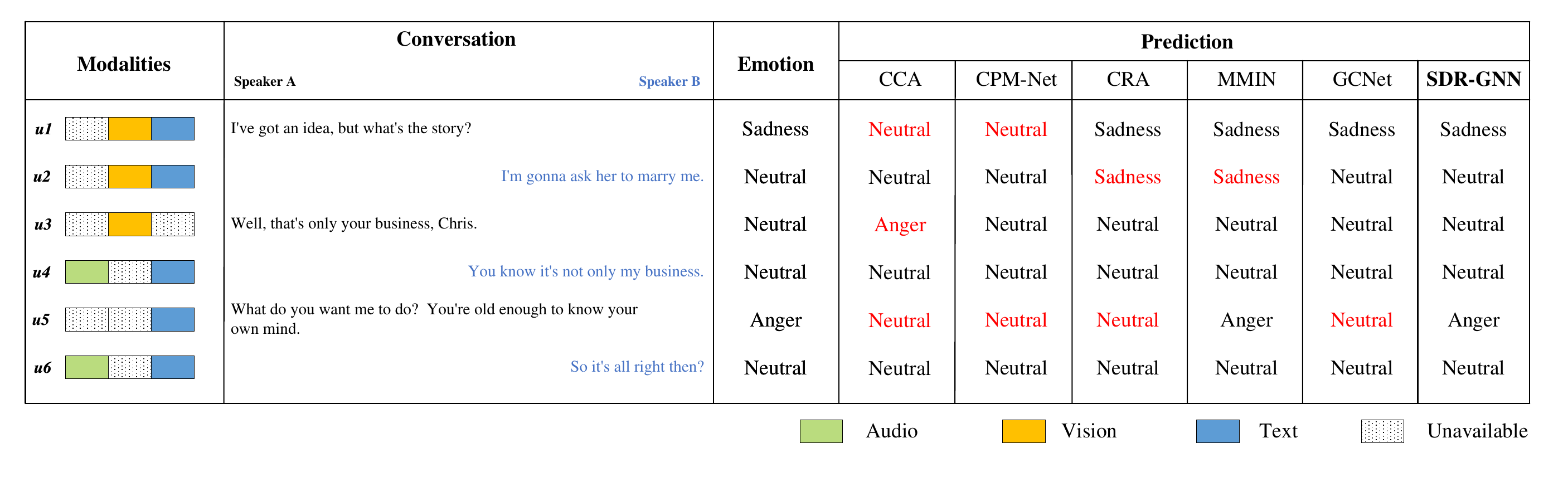}
	\caption{\textcolor{black}{Prediction results on incomplete conversational data from IEMOCAP (Four-class).}}
	\label{Figure8}
\end{figure*}

Our SDR-GNN model includes \textcolor{black}{four hyper-parameters}: the interaction window size $w$, the hidden layer dimension $h$, \textcolor{black}{reconstruction loss function weight $e$ and the number of hypergraphs layer $l$.} We assessed the impact of these parameters through experiments on the IEMOCAP (four-class) dataset under various missing rates, selecting $w$ from $\{1, 2, 3, 4\}$ and $h$ from $\{100,\textcolor{black}{150}, 200,\textcolor{black}{250}\}$, \textcolor{black}{$e$ ranges from $0.1$ to $0.9$ and $l$ from  $\{1, 2, 3, 4, 5, 6\}$.
The results of the experiments are displayed in Table \ref{Table5}, Fig. \ref{Figure7} and Fig. \ref{Figure8}.}

In most cases, the classification performance improves first the degrades as $w$ increase. This can be explained from two aspects. On one hand, a bigger window size can contain more utterances, which helps capturing and learning contextual information. On the other hand, a large window size will contain a large number of edges, which may include more irrelevant information. This will increase the difficulty of model learning.

\textcolor{black}{Similarly, an increase in the hidden layer dimension $h$ generally results in improved performance. At the same time, we also observed that when $h$ becomes too large, it leads to a decline in the model's performance. as seen in Table \ref{Table5}.} A larger $h$ provides a greater number of trainable parameters, thereby enhancing the model's ability to capture and represent complex feature interactions. This is particularly beneficial for discerning subtle patterns and distinctions in the data, which are crucial for accurate classification. However, this increase in parameters also heightens the risk of overfitting. Therefore, choosing appropriate parameters is crucial for improving the performance of the model.

To investigate the impact of the hyperparameters $e$ and $l$, we conducted experiments on the IEMOCAP (Four-class) dataset with a missing rate of $\mathcal{M} = 0.4$.

\textcolor{black}{As shown in Fig. \ref{Figure6}, when the weight of the reconstruction loss function $e$ increases from 0.1 to 0.9, the model's performance first rises and then declines, with the best performance observed around $e = 0.5$. We believe that the reconstruction task and the classification task should have similar weights. If the reconstruction task dominates, the classification results deteriorate; conversely, if the model focuses too much on the classification task, the quality of the reconstructed data decreases, which negatively impacts classification performance. Therefore, in our actual experiments, we set $e$ to 0.5 to balance the weights between the classification and reconstruction tasks, achieving good results in most cases.}

\textcolor{black}{From Fig. \ref{Figure7}, we can observe that as the number of layers $l$ increases, the model's performance also first improves and then declines. This is because with fewer layers, the number of propagated nodes is limited, while with more layers, redundant data propagation does not further enhance feature extraction. Additionally, it increases the risk of over-fitting.}

\textcolor{black}{\subsection{Case Study}}
\textcolor{black}{In this section, we compare the prediction results of different methods under the condition of missing modalities. The dialogue example is taken from IEMOCAP (Four-class), and Fig. \ref{Figure8} shows the experimental results. In this dialogue, Speaker B tells Speaker A that he intends to propose to Annie. We observe that as the degree of missing modalities increases, the performance of all models decreases, as it becomes more challenging to predict the outcome with less data.}

\textcolor{black}{In the dialogue shown in Fig. \ref{Figure8}, $u5$ can be considered a high-frequency signal sentence because SpeakerA shifts from Neutral to Anger, displaying intense emotion that stands out distinctly from the surrounding context. Most models perform poorly in predicting $u5$, except for SDR-GNN and MMIN. MMIN is designed to analyze individual utterances, so the surrounding context does not influence its predictions. In contrast, other models that rely on context for feature extraction or clustering algorithms may lose $u$5's high-frequency signal during the feature capturing and signal propagation process. However, SDR-GNN effectively differentiates multi-frequency information for feature aggregation, preventing this issue.}

\textcolor{black}{Our method consistently achieves high accuracy across all situations, demonstrating its effectiveness. SDR-GNN comprehensively leverages multi-frequency information, further improving prediction accuracy, which also highlights the importance of multi-frequency information.}

\section{Conclusion}
In this study, we introduce a novel framework, SDR-GNN, designed for addressing the challenges of incomplete multimodal learning in conversational emotion recognition. This approach leverages the dependencies between speakers and contexts, utilizing multi-frequency information within conversations effectively. Our framework specifically addresses the higher-order information of modalities and exploits multi-frequency data, bridging the gap in existing methods. We validate our method through experiments on three benchmark datasets, with results showing that SDR-GNN outperforms current methods in handling incomplete multimodal data for emotion recognition. Additionally, we dissect the critical role of each component within SDR-GNN and examine the influence of various hyper-parameters. Furthermore, After that, we analyze emotion categories at various missing rates and show the importance of incomplete data.

In the future, we will explore ways to better use the multi-frequency information in conversations and the relationships between the frequency signals and emotions.

\section*{CRediT authorship contribution statement}
\textbf{Fangze Fu}: Conceptualization, Methodology, Investigation, Data curation, Writing - Original Draft. \textbf{Wei Ai}: Supervision, Investigation, Writing - Review \& Editing. \textbf{Fang Yang}: Supervision, Writing - Review \& Editing. \textbf{Yuntao Shou}: Supervision, Investigation \& Review. \textbf{Tao Meng}: Supervision, Investigation, Writing - Review \& Editing. \textbf{Keqin Li}: Supervision, Investigation, Writing - Review \& Editing.

\section*{Declaration of Competing Interest}
The authors declare that they have no known competing financial interests or personal relationships that could have appeared to influence the work reported in this paper.

\section*{Data availability}
Data will be made available on request.

\section*{Acknowledgements}
\textcolor{black} {The authors deepest gratitude goes to the anonymous reviewers and AE for their careful work and thoughtful suggestions that have helped improve this paper substantially.} This work is supported by National Natural Science Foundation of China (Grant No. 69189338), Excellent Young Scholars of Hunan Province of China (Grant No. 22B0275), and program of Research on Local Community Structure Detection Algorithms in Complex Networks (Grant No. 2020YJ009).

\bibliographystyle{cas-model2-names}

\bibliography{reference}


\end{document}